\documentclass{article}

\usepackage[nonatbib,preprint]{neurips_2020}

\usepackage[utf8]{inputenc} % allow utf-8 input
\usepackage[T1]{fontenc}    % use 8-bit T1 fonts
\usepackage{hyperref}       % hyperlinks
\usepackage{url}            % simple URL typesetting
\usepackage{booktabs}       % professional-quality tables
\usepackage{amsfonts}       % blackboard math symbols
\usepackage{nicefrac}       % compact symbols for 1/2, etc.
\usepackage{microtype}      % microtypography

\usepackage{caption}        % style figure captions
\usepackage{booktabs}       % nicer looking tables
\usepackage{amsmath}        % math environments
\usepackage{multirow}       % multirow columns in tables
\usepackage{algorithm}      % algorithm figure environment
\usepackage{algpseudocode}  % pseudocode environment
\usepackage{listings}       % code listing package
\usepackage{enumitem}       % descrption lists, etc.
\usepackage{graphicx}       % include graphics
\usepackage{xcolor}         % color highlighting
\usepackage{float}          % more control over floats
\usepackage{subcaption}     % subfigures and subcaptions

\lstdefinelanguage{PDDL} {
    alsoletter={?,:,<,>,+,=,-,*,/},
    morecomment=[l]{\%},
    morekeywords={define, domain, problem},
    keywordsprefix={?},
    morekeywords=[2]{:requirements, :types, :constants,
                     :predicates, :derived, :functions,
                     :action, :parameters, :precondition, :effect, :objects, :init, :goal, :metric},
    keywords=[3]{and, not, forall, exists, imply,
                 -, +, *, /, <, >, >=, <=, =},
}

\lstdefinelanguage{Julia} {
    alsoletter={!,:},
    morecomment=[l]{\#},
    morestring=[b]",
    morekeywords={function},
    morekeywords=[2]{function, for, if, elseif, else, while, end, return, in, ::},
    morekeywords=[3]{@gen},
}

\title{Online Bayesian Goal Inference for \\ Boundedly-Rational Planning Agents}

\author{
  Tan Zhi-Xuan, \quad Jordyn L. Mann, \quad Tom Silver \\
  \textbf{Joshua B. Tenenbaum,} \qquad \textbf{Vikash K. Mansinghka} \\
  Massachusetts Institute of Technology \\
  \texttt{\{xuan,jordynm,tslvr,jbt,vkm\}@mit.edu}
}

\begin{document}

\maketitle

\begin{abstract}
  People routinely infer the goals of others by observing their actions over time. Remarkably, we can do so even when those actions lead to failure, enabling us to assist others when we detect that they might not achieve their goals. How might we endow machines with similar capabilities? Here we present an architecture capable of inferring an agent’s goals online from both optimal and non-optimal sequences of actions. Our architecture models agents as boundedly-rational planners that interleave search with execution by replanning, thereby accounting for sub-optimal behavior. These models are specified as probabilistic programs, allowing us to represent and perform efficient Bayesian inference over an agent's goals and internal planning processes. To perform such inference, we develop Sequential Inverse Plan Search (SIPS), a sequential Monte Carlo algorithm that exploits the online replanning assumption of these models, limiting computation by incrementally extending inferred plans as new actions are observed. We present experiments showing that this modeling and inference architecture outperforms Bayesian inverse reinforcement learning baselines, accurately inferring goals from both optimal and non-optimal trajectories involving failure and back-tracking, while generalizing across domains with compositional structure and sparse rewards.
\end{abstract}

\section{Introduction}

Everyday experience tells us that it is impossible to plan ahead for everything. Yet, not only do humans still manage to achieve our goals by piecing together partial and approximate plans, we also appear to account for this cognitive strategy when inferring the goals of others, understanding that they might plan and act sub-optimally, or even fail to achieve their goals.  Indeed, even 18-month old infants seem capable of such inferences, offering their assistance to adults after observing them execute failed plans \cite{warneken2006altruistic}. How might we understand this ability to infer goals from such plans? And how might we endow machines with this capacity, so they might assist us when our plans fail?

While there has been considerable work on inferring the goals and desires of agents, much of this work has assumed that agents act optimally to achieve their goals. Even when this assumption is relaxed, the forms of sub-optimality considered are often highly simplified. In inverse reinforcement learning, for example, agents are assumed to either act optimally \cite{ng2000algorithms} or to exhibit Boltzmann-rational action noise \cite{ziebart2008maximum}, while in the plan recognition literature, longer plans are assigned exponentially decreasing probability \cite{ramirez2010probabilistic}. None of these approaches account for the difficulty of planning itself, which may lead agents to produce sub-optimal or failed plans. This not only makes them ill-equipped to infer goals from such plans, but also saddles them with a cognitively implausible burden: If inferring an agent's goals requires knowing the optimal solution to reach each goal, then an observer would need to compute the optimal plan or policy for \emph{all} of those goals in advance \cite{michini2012improving}. Outside of the simplest problems and domains, this is deeply intractable.

\begin{figure}[t]
    \centering
    \includegraphics[width=\textwidth]{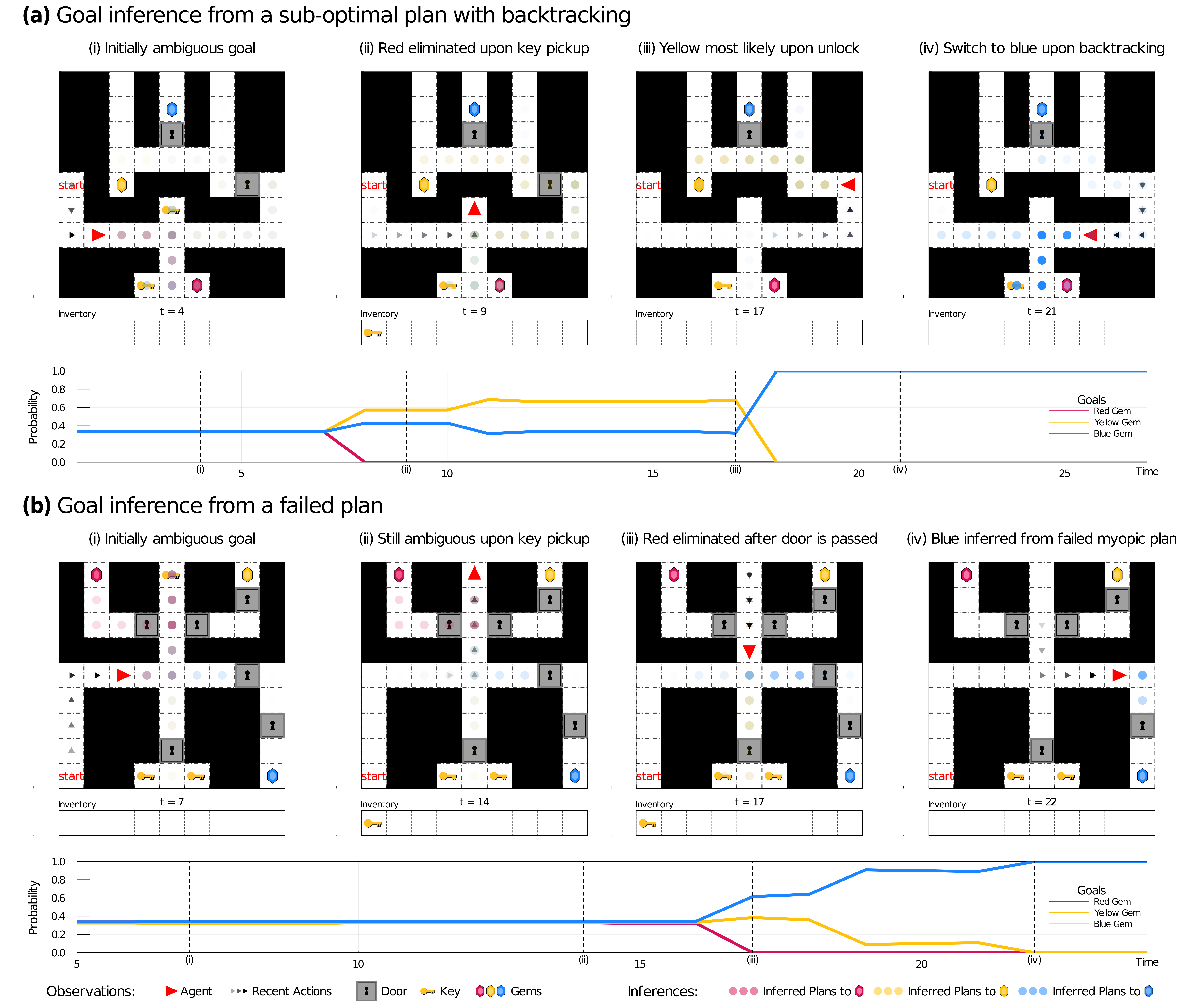}
    \caption{Our architecture performing online Bayesian goal inference via Sequential Inverse Plan Search. In \textbf{(a)}, an agent exhibits a \emph{sub-optimal plan} to acquire the blue gem, backtracking to pick up the key required for the second door. In \textbf{(b)}, an agent exhibits a \emph{failed plan} to acquire the blue gem, myopically using up its first key to get closer to the gem instead of realizing that it needs to collect the bottom two keys. In both cases, our method not only manages to infer the correct goal by the end, but also captures sharp human-like shifts in its inferences at key points, such as \textbf{(a.ii)} when the agent picks up a key unnecessary for the red gem, \textbf{(a.ii)} when the agent starts to backtrack, \textbf{(b.iii)} when the agent ignores the door to the red gem, or \textbf{(b.iv)} when the agent unlocks the first door to the blue gem. }
    \label{fig:goal-inference-storyboards}
\end{figure}

\begin{figure}
    \centering
    \includegraphics[width=\textwidth]{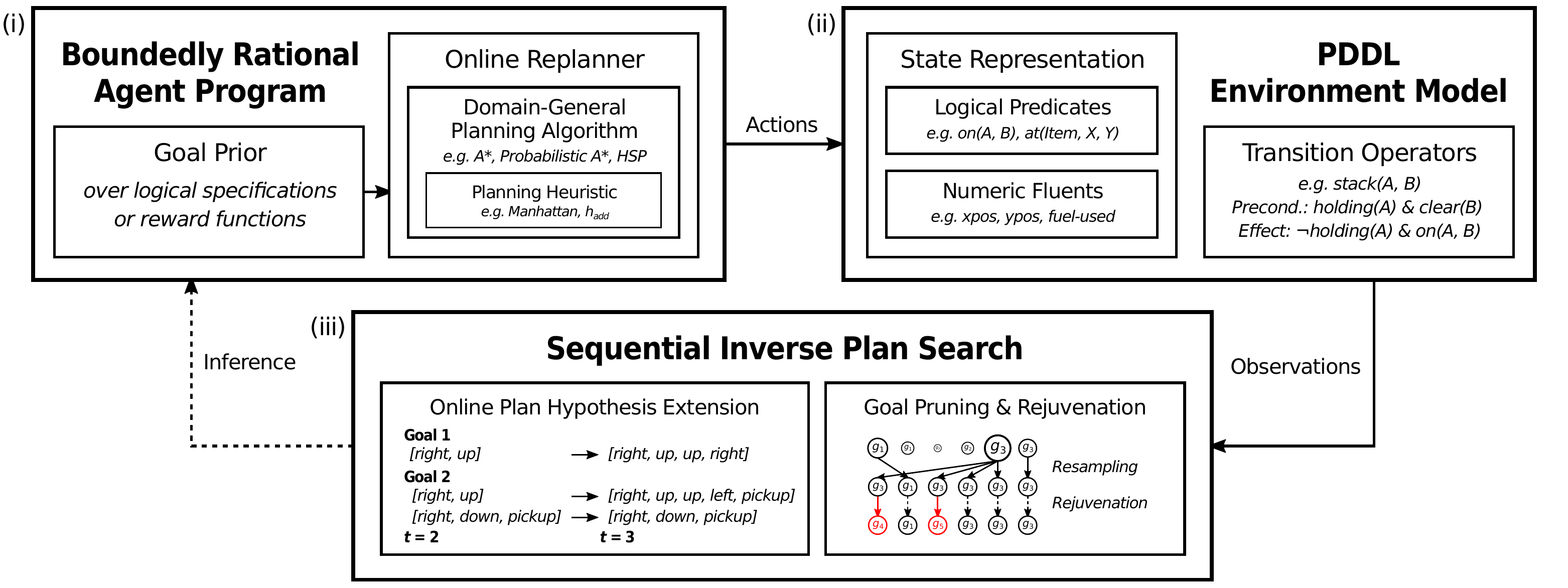}
    \caption{Our modeling and inference architecture is comprised of: \textbf{(i)} A programmatic model of a boundedly rational planning agent, implemented in the Gen probabilistic programming system; \textbf{(ii)} An environment model specified in the Planning Domain Definition Language (PDDL), facilitating support for a wide variety of planning domains and state-of-the-art symbolic planners; \textbf{(iii)} Sequential Inverse Plan Search (SIPS), a novel SMC algorithm that exploits the replanning assumption of our agent model to reduce computation, extending hypothesized plans only as new observations arrive.}
    \label{fig:architecture-overview}
\end{figure}

In this paper, we present a unified modeling and inference architecture (Figure \ref{fig:architecture-overview}) that addresses both of these limitations. In contrast to prior work that models agents as \emph{actors} that are \emph{noisily rational}, we model agents as \emph{planners} that are \emph{boundedly rational} with respect to how much they plan, interleaving resource-limited plan search with plan execution. This allows us to perform online Bayesian inference of plans and goals even from highly sub-optimal trajectories involving backtracking or irreversible failure (Figure \ref{fig:goal-inference-storyboards}). We do so by modeling agents as probabilistic programs (Figure \ref{fig:agent-model}), comprised of goal priors and domain-general planning algorithms (Figure \ref{fig:architecture-overview}i), and interacting with a symbolic environment model (Figure \ref{fig:architecture-overview}ii). Inference is then performed via Sequential Inverse Plan Search (SIPS), a sequential Monte Carlo (SMC) algorithm that exploits the replanning assumption of our agent models, incrementally inferring partial plans while limiting computational cost (Figure \ref{fig:architecture-overview}iii).

Our architecture delivers both accuracy and speed by being built in Gen, a general-purpose probabilistic programming system that supports customized inference using data-driven proposals and involutive rejuvenation kernels \cite{cusumano2019gen,cusumano2018using,cusumano2020automating}, alongside an embedding of the Planning Domain Definition Language \cite{mcdermott1998pddl,fox2003pddl2}, enabling the use of fast general-purpose planners \cite{bonet2001planning} as modeling components. We evaluate our approach against a Bayesian inverse reinforcement learning baseline \cite{ramachandran2007bayesian} on a wide variety of planning domains that exhibit compositional task structure and sparse rewards (e.g. Figure \ref{fig:goal-inference-storyboards}), achieving high accuracy on many domains, often with orders of magnitude less computation.

\section{Related Work}

\textbf{Inverse reinforcement learning (IRL).} A long line of work has shown how to learn reward functions as explanations of goal-directed agent behavior via inverse reinforcement learning \cite{ng2000algorithms, abbeel2004apprenticeship, ramachandran2007bayesian,hadfield2016cooperative}. However, most such approaches are too costly for online settings of complex domains, as they require solving the underlying Markov Decision Process (MDP) for every posited goal or reward function, and for all possible initial states \cite{brown2019deep, michini2012improving}. Our approach instead assumes that agents are online model-based planners. This greatly reduces computation time, while also better reflecting humans' intuitive understanding of other agents.

\textbf{Bayesian theory-of-mind (BToM).} Computational models of humans' intuitive theory-of-mind posit that we understand other's actions by Bayesian inference of their likely goals and beliefs. These models, largely built upon the same MDP formalism used in IRL, have been shown to make predictions that correspond closely with human inferences \cite{goodman2006intuitive,baker2007goal,baker2009action,baker2011bayesian,jara2016naive,baker2017rational,jara2019naive}. Some recent work also models agents using probabilistic programs \cite{cusumano2017probabilistic,seaman2018nested}. Our research extends this line of work by explicitly modeling an agent's partial plans, or \emph{intentions} \cite{bratman1987intention}. This allows our architecture to infer final goals from instrumental subgoals produced as part of a plan, and to account for sub-optimality in those plans, thereby enriching the range of mental inferences that BToM models can explain.

\textbf{Plan recognition as planning (PRP).} Our work is related to the literature on plan recognition as planning, which performs goal and plan inference by using classical satisficing planners to model plan likelihoods given a goal \cite{ramirez2009plan,ramirez2010probabilistic,sohrabi2016plan,holler2018plan,kaminka2018plan,vered2018online}. However, because these approaches use a heuristic likelihood model that assumes goals are always achievable, they are unable to infer likely goals when irreversible failures occur. In contrast, we model agents as online planners who may occasionally execute partial plans that lead to dead ends.

\textbf{Online goal inference.} Several recent papers have extended IRL to an online setting, but these have either focused on maximum-likelihood estimation in 1D state spaces \cite{thaker2017online,self2019online}, or utilize an expensive value iteration subroutine that is unlikely to scale \cite{rhinehart2018first}. In contrast, we develop a sequential Monte Carlo algorithm that exploits the online nature of the agent models in order to perform incremental plan inference with limited computation cost.

\textbf{Inferences from sub-optimal behavior.} We build upon a growing body of research on inferring goals and preferences while accounting for human sub-optimality \cite{ziebart2008maximum,  seaman2018nested, evans2015learning, evans2016learning, shah2019feasibility, armstrong2018occam}, introducing a model of boundedly-rational planning as resource-limited search. This reflects a natural principle of \emph{resource rationality} under which agents are less likely to engage in costly computations \cite{griffiths2015rational,ho2020efficiency}. Unlike prior models of myopic agents which assign zero reward to future states beyond some time horizon \cite{evans2015learning,shah2019feasibility}, our approach accounts for myopic planning in domains with instrumental subgoals and sparse rewards.

\section{Boundedly-Rational Planning Agents}

In order to account for sub-optimal behavior due to resource-limited planning, observers need to model not only an agent's goals and actions, but also the plans they form to achieve those goals. As such, we model agents and their environments as generative processes of the following form:
\begin{alignat}{2}
\textit{Goal prior:}& \qquad g &&\sim P(g) \\
\textit{Plan update:}& \qquad p_t &&\sim P(p_t | s_t, p_{t-1}, g) \\
\textit{Action selection:}& \qquad a_t &&\sim P(a_t | s_t, p_t) \\
\textit{State transition:}& \quad s_{t+1} &&\sim P(s_{t+1} | s_t, a_t) \\
\textit{Observation noise:}& \quad o_{t+1} &&\sim P(o_{t+1}|s_{t+1})
\end{alignat}
where $g$, $p_t$, $a_t$, $s_t$ are the agent's goals, the internal state of the agent's plan, the agent's action, and the environment's state at time $t$ respectively. For the purposes of goal inference, observers also assume that each state $s_t$ might be subject to observation noise, producing an observed state $o_t$.

\begin{figure}[t]
\centering
\begin{subfigure}[b]{0.54\textwidth}
\includegraphics[width=\textwidth]{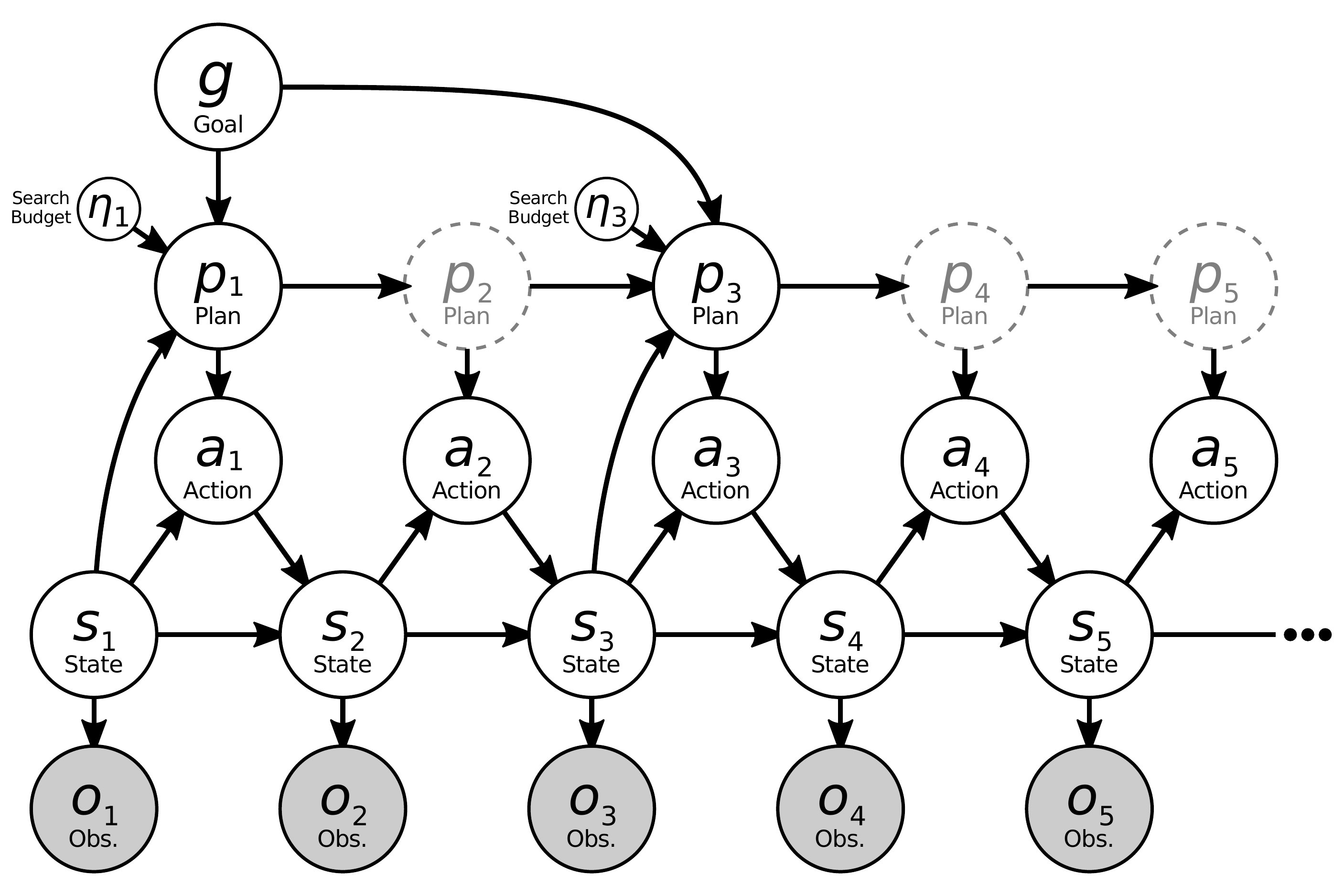}
\caption{One realization of our agent and environment model.}
\label{fig:graphical-model}
\end{subfigure}
\hspace{0.03\textwidth}
\begin{subfigure}[b]{0.4\textwidth}
\centering
\algrenewcommand\algorithmicprocedure{\textbf{model}}
\begin{algorithmic}
\scriptsize
\Procedure{update-plan}{$t$, $s_t$, $p_{t-1}$, $g$}
\State \textbf{parameters}: \textsc{planner}, $r, q, \gamma, h$
\If{$t > \textsc{length}(p_{t-1})$ \textbf{or} $s_t \notin p_{t-1}[t]$}
    \State $\eta \sim \textsc{negative-binomial}(r, q)$
    \State $\tilde{p}_t \sim \textsc{planner}(s_t, g, h, \gamma, \eta)$
    \State $p_t \gets \textsc{append}(p_{t-1}, \tilde{p}_t)$
\Else
    \State $p_t \gets p_{t-1}$
\EndIf
\State \Return $p_t$
\EndProcedure
\end{algorithmic}
\scriptsize{\textbf{(i)} Samples from $P(p_t | s_t, p_{t-1}, g)$}
\vspace{3pt}
\begin{algorithmic}
\scriptsize
\Procedure{select-action}{$t$, $s_t$, $p_t$}      \State \Return $p_t[t][s_t]$
\EndProcedure
\end{algorithmic}
\scriptsize{\textbf{(ii)} Samples from $P(a_t | s_t, p_t)$}
\vspace{1pt}
\caption{Boundedly-rational agent programs.}
\label{fig:agent-subroutines}
\end{subfigure}
\caption{We model agents as boundedly rational planners that interleave search and execution of partial plans as they interact with the environment. In \textbf{(a)} we depict one possible realization of this model, where the agent initially samples a search budget $\eta_1$ and searches for a plan $p_1$ that is two actions long. At $t=2$, no additional planning needs to be done, so $p_2$ is copied from $p_1$, as denoted by the dashed lines. The agent then replans at $t=3$ from state $s_3$, sampling a new search budget $\eta_3$ and an extended plan $p_3$ with three more actions. We formally specify this agent model using probabilistic programs, with pseudo-code shown in \textbf{(b)}. \textsc{update-plan} samples extended plans $p_t$ given previous plans $p_{t-1}$, while \textsc{select-action} selects an action $a_t$ according the current plan $p_t$.}
\label{fig:agent-model}
\end{figure}

This generative process, depicted in Figure \ref{fig:agent-model}a, extends the standard model of MDP agents by modeling plans and plan updates explicitly, allowing us to represent not only agents that act according to some precomputed policy $a_t \sim \pi(a_t|s_t)$, but also agents that compute and update their plans $p_t$ on-the-fly. We describe each component of this process in greater detail below.

\subsection{Modeling Goals, States and Observations}

To represent states, observations, goals, and distributions over goals in a general and flexible manner, our architecture embeds the Planning Domain Definition Language (PDDL) \cite{mcdermott1998pddl,fox2003pddl2}, representing states $s_t$ and goals $g$ in terms of predicate-based facts, relations, and numeric expressions (Figure \ref{fig:architecture-overview}ii). State transitions $P(s_t | s_{t-1}, a_{t-1})$ are modeled by transition operators that specify the preconditions and effects of actions. While we focus on deterministic transitions in this paper, we also support stochastic transitions, as in Probabilistic PDDL \cite{younes2004ppddl1}. Given this representation, an observer's prior over goals $P(g)$ can be specified as a probabilistic program over PDDL goal specifications, including numeric reward functions, as well as sets of goal predicates (e.g. \texttt{has(gem)}), equivalent to indicator reward functions. Observation noise $P(o_{t+1}|s_{t+1})$ can also be modeled by corrupting each Boolean predicate with some probability, and adding continuous (e.g. Gaussian) noise to numeric fluents.

\subsection{Modeling Sub-Optimal Plans and Actions}

To model sub-optimal plans, the basic insight we follow is that agents like ourselves are \emph{boundedly rational}: we \emph{attempt} to plan to achieve our goals efficiently, but are limited by our cognitive resources. The primary limitation we consider is that full-horizon planning is often costly or intractable. Instead, it may often make sense to form partial plans towards promising intermediate states, execute them, and replan from there. We model this by assuming that agents only search for a plan up to some budget $\eta$, before executing a partial plan to a promising state found during search. We operationalize $\eta$ as the maximum number of nodes expanded (i.e., states explored), which we treat as a random variable sampled from a negative binomial distribution:
\begin{equation}
    \eta \sim \textsc{negative-binomial}(r, q)
\end{equation}
The parameters $r$ (maximum failure count) and $q$ (continuation probability) characterize the persistence of a planner who may choose to give up after expanding each node. When $r > 1$, this distribution peaks at medium values of $\eta$, then decreases exponentially, modeling agents that are unlikely to form extremely long plans, which are costly, or extremely short plans, which are unhelpful.

This model also assumes access to a planning algorithm capable of producing partial plans. While we support \emph{any} such planner as a sub-component, in this work we focus on A* search due to its ability to support domain-general heuristics that can guide search in human-like ways \cite{bonet2001planning,geffner2010heuristics}. We also modify A* so that search is stochastic, modeling agent sub-optimality during search. In particular, instead of always expanding the most promising successor state, we sample successor $s$ with probability:
\begin{equation}
   P_\text{expand}(s) \propto \exp(-f(s, g)/\gamma)
\end{equation}
where $\gamma$ is a noise parameter controlling the randomness of search, and $f(s, g) = c(s) + h(s, g)$ is the estimated total plan cost, i.e. the sum of the path cost $c(s)$ so far with the heuristic goal distance $h(s, g)$. On termination, we simply return the most recently selected successor state, which is likely to have low total plan cost $f(s, g)$ if the heuristic  $h(s, g)$ is informative and the noise $\gamma$ is low.

We incorporate these limitations into a model of how a boundedly rational planning agent interleaves search and execution, specified by the probabilistic programs \textsc{update-plan} and \textsc{select-action} in Figure \ref{fig:agent-model}b. At each time $t$, the agent may reach the end of its last made plan $p_{t-1}$ or encounter a state $s_t$ not anticipated by the plan, in which case it will call the base planner (probabilistic A*) with a randomly sampled node budget $\eta$. The partial plan produced is then used to extend the original plan. Otherwise, the agent will simply continue executing its original plan, performing no additional computation. Note that by replanning when the unexpected occurs, the agent automatically handles some amount of stochasticity, as well as errors in its environment model.

\section{Online Bayesian Goal Inference}

Having specified our model, we can now state the problem of Bayesian goal inference. We assume that an observer receives a sequence of potentially noisy state observations $o_{1:t} = (o_1, ..., o_t)$. Given the observations up to timestep $t$ and a set of possible goals $\mathcal{G}$, the observer's aim is to infer the agent's goal $g \in \mathcal{G}$ by computing the posterior:
\begin{equation}
    P(g|o_{1:t}) \propto P(g) \textstyle\sum_{\substack{\tiny s_{1:t}\\a_{1:t}\\p_{1:t}}} \prod_{\tau=0}^{t-1} P(o_{\tau+1}|s_{\tau+1}) P(s_{\tau+1} | s_\tau, a_\tau) P(a_\tau | s_\tau, p_\tau) P(p_\tau | s_\tau, p_{\tau-1}, g)
\end{equation}

Computing this posterior exactly is intractable, as it requires marginalizing over all the random latent variables $s_\tau$, $a_\tau$, and $p_\tau$. Instead, we develop a sequential Monte Carlo procedure, shown in Algorithm \ref{alg:sips}, to perform approximate inference in an online manner, using samples from the posterior $P(g|o_{1:t-1})$ at time $t-1$ to inform sampling from the posterior $P(g|o_{1:t})$ at time $t$. We call this algorithm Sequential Inverse Plan Search (SIPS), because it sequentially inverts a search-based planning algorithm, inferring sequences of partial plans that are likely given the observations, and consequently the likely goals.

As in standard particle filtering schemes, we first sample a set of particles or hypotheses $i \in [1,k]$, with corresponding weights $w_i$ (lines 3-5). Each particle corresponds to a particular plan $p^i_\tau$ and goal $g^i$. As each new observation $o_\tau$ arrives, we extend the particles (lines 12--14) and reweight them by their likelihood of producing that observation (line 15). The collection of weighted particles thus approximates the full posterior over the unobserved variables in our model, including the agent's plans and goals. We describe several key features of this algorithm below.

\begin{algorithm}[t]
\caption{Sequential Inverse Plan Search (SIPS) for online Bayesian goal inference}
%\scriptsize
\fontsize{8pt}{8pt}\selectfont
\label{alg:sips}
\begin{algorithmic}[1]
\Procedure{sips}{$s_0$, $o_{1:t}$, }
\State \textbf{parameters:} $k$, number of particles; $c$, resampling threshold
\State $w^i \gets 1$ \textbf{for} $i \in [1, k]$ \Comment{Initialize particle weights}
\State $s_0^i, p_0^i, a_0^i \gets s_0, [], \text{no-op}$  \textbf{for} $i \in [1, k]$ \Comment{Initialize states, plans and actions}
\State $g^i \sim \textsc{goal-prior}()$ \textbf{for} $i \in [1, k]$ \Comment{Sample $k$ particles from goal prior}
\For{$\tau \in [1,t]$}
    \If{$\textsc{effective-sample-size}(w^1, ..., w^k)/k < c$} \Comment{Resample and rejuvenate}
        \State $g^i, s_{1:\tau}^i, p_{1:\tau}^i, a_{1:\tau}^i \sim \textsc{resample}([g^i, s_{1:\tau}, p_{1:\tau}, a_{1:\tau}]^{1:k})$ \textbf{for} $i \in [1, k]$
        \State $g^i, s_{1:\tau}^i, p_{1:\tau}^i, a_{1:\tau}^i \sim \textsc{rejuvenate}(g^i, o_{1:\tau}, s_{1:\tau}^i, p_{1:\tau}^i, a_{1:\tau}^i)$ \textbf{for} $i \in [1, k]$
    \EndIf
    \For{$i \in [1, k]$} \Comment{Extend each particle to timestep $\tau$}
        \State $s_\tau^i \sim P(s_{\tau} | s_{\tau-1}^i, a_{\tau-1}^i)$ \Comment{Sample state transition}
        \State $p_\tau^i \sim \textsc{update-plan}(p_\tau | s_\tau^i, p_{\tau-1}^i,  g^i)$ \Comment{Extend plan if necessary}
        \State $a_\tau^i \sim \textsc{select-action}(a_\tau | s_\tau^i, p_\tau^i)$ \Comment{Select action}
        \State $w^i \gets w^i \cdot P(o_{\tau} | s_{\tau}^i)$ \Comment{Update particle weight}
    \EndFor
\EndFor
\State $\tilde w^i \gets w^i / \sum_{j=1}^k w^j$ \textbf{for} $i \in [1, k]$ \Comment{Normalize particle weights}
\State \Return $[(g^1, w^1), ..., (g^k, w^k)]$ \Comment{Return weighted goal particles}
\EndProcedure
\State
\Procedure{rejuvenate}{$g$, $o_{1:\tau}$, $s_{1:\tau}$, $p_{1:\tau}$, $a_{1:\tau}$} \Comment{Metropolis-Hasting rejuvenation move}
    \State \textbf{parameters:} $p_g$, goal rejuvenation probability
    \If{\textsc{bernoulli}($p_g$)} \Comment{Heuristic-driven goal proposal}
        \State $g' \sim Q(g) := \textsc{softmax}$([$h(o_\tau, g)$ for $g \in \mathcal{G}$]) \Comment{Propose $g'_0$ based on est. distance to $o_\tau$}
        \State $s'_{1:\tau}, p'_{1:\tau}, a'_{1:\tau} \sim P(s_{1:\tau}, p_{1:\tau}, a_{1:\tau}|g)$  \Comment{Sample trajectory under new goal $g$}
        \State $\alpha \gets Q(g) / Q(g')$ \Comment{Compute proposal ratio}
    \Else  \Comment{Error-driven replanning proposal}
        \State $t_* \sim Q(t_*|s_{1:\tau}, o_{1:\tau})$ \Comment{Sample a time close to when $s_{1:\tau}$ diverges from $o_{1:\tau}$}
        \State $s'_{t_*:\tau}, p'_{t_*:\tau}, a'_{t_*:\tau} \sim Q(s_{t_*:\tau}, p_{t_*:\tau}, a_{t_*:\tau}|o_{t_*:\tau})$ \Comment{Propose new plan sequence $p'_{t_*:\tau}$}
        \State $\alpha \gets Q(s_{t_*:\tau}, p_{t_*:\tau}, a_{t_*:\tau}|o_{t_*:\tau}) / Q(s'_{t_*:\tau}, p'_{t_*:\tau}, a'_{t_*:\tau}|o_{t_*:\tau})$ \Comment{Compute proposal ratio}
        \State $\alpha \gets \alpha \cdot Q(t_*|s'_{1:\tau}, o_{1:\tau}) / Q(t_*|s_{1:\tau}, o_{1:\tau})$ \Comment{Reweight by auxiliary proposal ratio}
    \EndIf
    \State $\alpha \gets \alpha \cdot P(o_{1:\tau}|s'_{1:\tau}) / P(o_{1:\tau}|s_{1:\tau})$ \Comment{Compute acceptance ratio}
    \State \Return $g'_0, s'_{1:\tau}, p'_{1:\tau}, a'_{1:\tau}$ \textbf{if} \textsc{bernoulli}($\min(\alpha, 1)$) \textbf{else} $g_0, s_{1:\tau}, p_{1:\tau}, a_{1:\tau}$ \Comment{Accept or reject proposals}
\EndProcedure
\end{algorithmic}
\end{algorithm}

\subsection{Online Extension of Hypothesized Partial Plans}

A key aspect that makes SIPS a genuinely \emph{online} algorithm is the modeling assumption that agents also plan \emph{online}. This obviates the need for the observer to precompute a complete plan or policy for each of the agent's possible goals in advance, and instead defers such computation to the point where the agent reaches a time $t$ that the observer's hypothesized plans do not yet reach. In particular, for each particle $i$, the corresponding plan hypothesis $p^i_{t-1}$ is extended (Algorithm \ref{alg:sips}, line 13) by running the \textsc{update-plan} procedure in Figure \ref{fig:agent-model}b.i, which only performs additional computation if $p^i_{t-1}$ does not already contain a planned action for time $t$ and state $s_t$. This means that at any given time $t$, only a small number of plans require extension, limiting the number of expensive planning calls.

\subsection{Managing Hypothesis Diversity via Resampling and Rejuvenation}

We also introduce resampling and rejuvenation steps into SIPS in order to ensure particle diversity. Whenever the effective sample size falls below a threshold $c$ (line 7), we resample the particles (line 8), thereby pruning low-weight hypotheses. We then rejuvenate by applying a mixture of two data-driven Metropolis-Hastings kernels to each particle. The first kernel uses a heuristic-driven goal proposal (lines 25-27), proposing goals $\tilde g \in \mathcal{G}$ which are close in heuristic distance $h(o_\tau, \tilde g)$ to the last observed state $o_\tau$. This allows SIPS to reintroduce goals that were pruned, but later become more likely. The second kernel uses an error-driven replanning proposal (lines 29-32), which samples a time close to the divergence point between the hypothesized and observed trajectories, and then proposes to replan from that time, thereby constructing a new sequence of hypothesized partial plans that are less likely to diverge from the observations. Despite the complexity of these proposals, acceptance ratios are automatically calculated via Gen's support for involutive kernels \cite{cusumano2020automating}. Collectively, these steps help to ensure that hypotheses are both diverse and likely given the observations.

\section{Experiments}

We conducted several sets of experiments that demonstrate the human-likeness, accuracy, speed, and robustness of our approach. We first present experiments demonstrating the novel capacity of SIPS to infer goals from sub-optimal trajectories involving backtracking and failure (Figure \ref{fig:goal-inference-storyboards}). Comparing these inferences against human goal inferences shows that SIPS is more human-like than baseline approaches (Figure \ref{fig:human-likeness}). We also evaluate the accuracy and speed of SIPS on a variety of planning domains (Table \ref{tab:performance}), showing that it outperforms Bayesian IRL baselines. Finally, we present robustness experiments showing that SIPS can infer goals even when the data-generating model differs from the model assumed by the algorithm (Table \ref{tab:robustness}).

\subsection{Domains}

We validate our approach on domains with varying degrees of complexity, both in terms of the size of the state space $|\mathcal{S}|$ and the number of possible goals $|\mathcal{G}|$. All domains are characterized by compositional structure and sparse rewards, posing a challenge for standard MDP-based approaches.

\textbf{Taxi} ($|\mathcal{G}|=3, |\mathcal{S}| = 125$): A benchmark domain used in hierarchical reinforcement learning \cite{dietterich1998maxq}, where a taxi has to transport a passenger from one location to another in a gridworld.

\textbf{Doors, Keys, \& Gems} ($|\mathcal{G}|=3, |\mathcal{S}| \sim 10^5$): A domain in which an agent must navigate a maze with doors, keys, and gems (Figure \ref{fig:goal-inference-storyboards}). Each key can be used once to unlock a door, allowing the agent to acquire items behind that door. Goals correspond to acquiring one out of three colored gems.

\textbf{Block Words} ($|\mathcal{G}|=5, |\mathcal{S}| \sim 10^5$): A Blocks World variant adapted from \cite{ramirez2010probabilistic} where blocks are labeled with letters. Goals correspond to block towers that spell one of a set of five English words.

\textbf{Intrusion Detection} ($|\mathcal{G}|=20, |\mathcal{S}| \sim 10^{30}$): A cybersecurity-inspired domain drawn from \cite{ramirez2010probabilistic}, where an agent might perform a variety of attacks on a set of  servers. There are 20 possible goals, each corresponding to a set of attacks (e.g. cyber-vandalism or data-theft) on up to 10 servers.

\subsection{Baselines}

We implemented Bayesian IRL (BIRL) baselines by running value iteration to compute a Boltzman-rational policy $\pi(a_t|s_t, g)$ for each possible goal $g \in \mathcal{G}$. Following the setting of early Bayesian theory-of-mind approaches \cite{baker2009action}, we treated goals as indicator reward functions, and assumed a uniform prior $P(g)$ over goals.  Inference was then performed by exact computation of the posterior over reward functions, using the policy as the likelihood for observed actions. Unless otherwise stated, we used a discount factor of 0.9, and Boltzmann noise parameter $\alpha$=$1$.

Due to the exponentially large state space of many of our domains, standard value iteration (VI) often failed to converge even after $10^6$ iterations. As such, we implemented two variants of BIRL that use asynchronous VI, sampling states instead of fully enumerating them. The first, \emph{unbiased BIRL}, uses uniform random sampling of the state space up to 250,000 iterations, sufficient for convergence in the Block Words domain. The second, \emph{oracle BIRL}, assumes oracular access to the full set of observed trajectories in advance, and performing biased sampling of states that appear in those trajectories. Although inapplicable in practice for online use, this ensures that the computed policy is able to reach the goal in all cases, making it a useful benchmark for comparison.

\subsection{Human-Like Goal Inference from Sub-optimal and Failed Plans}

\begin{figure}[t]
    \centering
    \includegraphics[width=\textwidth]{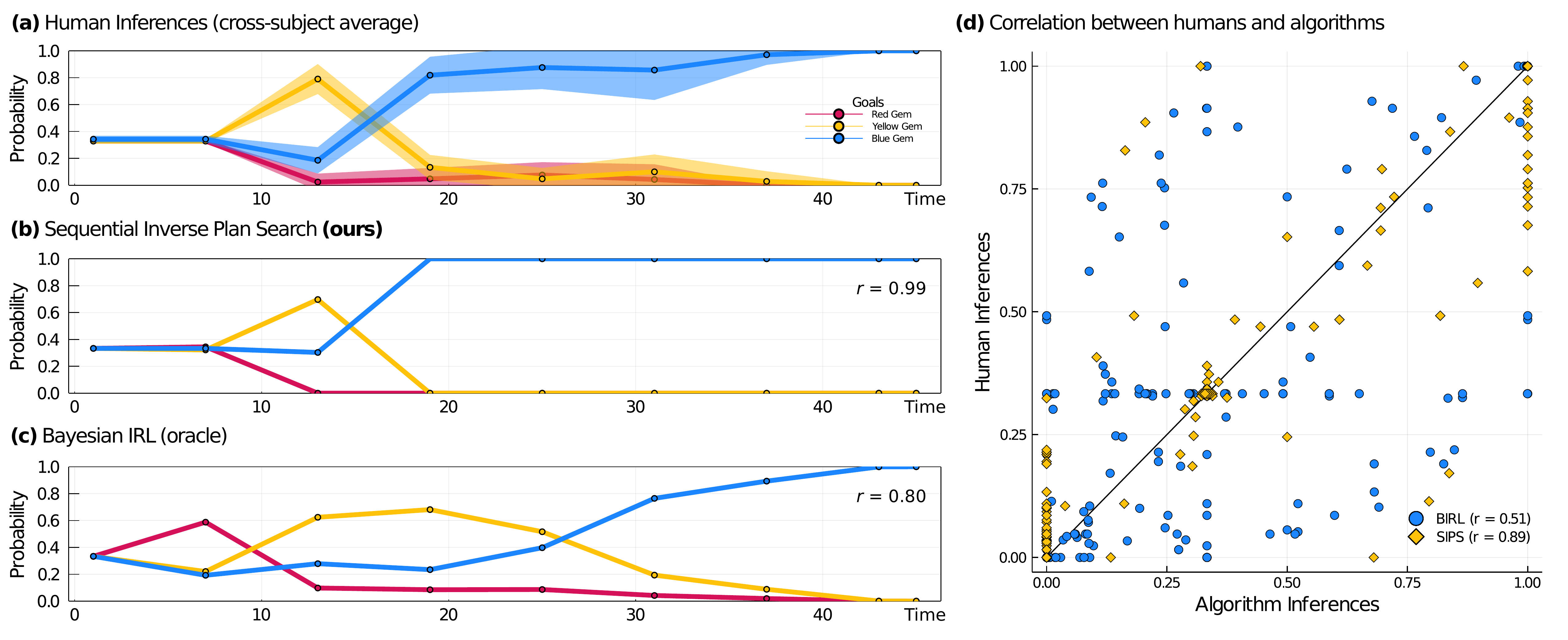}
    \caption{\textbf{(a)} Average human goal inferences over time ($\pm 1$ std.) for the sub-optimal trajectory in Figure \ref{fig:goal-inference-storyboards}a, compared to \textbf{(b)} inferences made by SIPS and \textbf{(c)} oracle BIRL. We omit unbiased BIRL because unbiased VI fails to converge for this domain, producing a flat posterior. In \textbf{(d)} we show a scatterplot of mean human inferences against algorithm inferences across all trajectories.}
    \label{fig:human-likeness}
\end{figure}

To investigate the novel human-like capabilities of our approach, we performed a set of qualitative experiments on a set of trajectories designed to exhibit notable sub-optimality or failure. The experiments were performed on the Doors, Keys \& Gems domain because it allows for irreversible failures. Two illustrative examples are shown in Figure \ref{fig:goal-inference-storyboards}, and more are provided in the supplement. In Figure \ref{fig:goal-inference-storyboards}a, SIPS accurately infers goals from a sub-optimal plan with backtracking, initially placing more posterior mass on the yellow gem when the agent acquires the first key (panel ii), but then switching to the blue gem once the agent backtracks to the second key (panel iv). In Figure \ref{fig:goal-inference-storyboards}b, SIPS remains uncertain about all three goals when the first key is acquired (panel ii), but discards the red gem as a possibility when the agent walks past the door (panel iii), and finally converges upon the blue gem when the agent myopically unlocks the first door required to access that gem (panel iv).

In addition to posterior convergence, the inferences made by SIPS display human-like changes at key timepoints. We quantified this human-likeness by collecting human goal inferences on ten trajectories (six sub-optimal or failed) in a pilot study with $N$=$8$ subjects. Human inferences were collected every six timesteps, and a comparison against SIPS and the oracle BIRL baseline is shown in Figure \ref{fig:human-likeness}. For the trajectory in in Figure \ref{fig:goal-inference-storyboards}a, human inferences (Figure \ref{fig:human-likeness}a) display extremely similar qualitative trends as SIPS (Figure \ref{fig:human-likeness}b, $r$=$0.99$). Oracle BIRL correlates less well (Figure \ref{fig:human-likeness}c, $r$=$0.80$), assigning high probability to the yellow gem even after the agent backtracks at $t \geq 18$. This is because Boltzmann action noise assigns significant likelihood to the undoing of past actions. As Figure \ref{fig:human-likeness}d shows, SIPS also correlates more strongly with mean human inferences across the dataset. Inferences made SIPS (yellow) hew closely to the diagonal, achieving a correlation of $r$=$0.89$, indicating that the agent model assumed by SIPS is similar to humans' theory-of-mind. In contrast, inferences made by BIRL (blue) are much more diffuse, achieving a correlation of only $r$=$0.51$.

\subsection{Accuracy, Speed and Robustness of Inference}

To evaluate accuracy and speed, we ran each inference method on a dataset of optimal and non-optimal agent trajectories for each domain, assuming a uniform prior over goals. The optimal trajectories were generated using A* search with an admissible heuristic for each possible goal in the domain. Non-optimal trajectories were generated using the replanning agent model in Figure \ref{fig:agent-model}b, with parameters $r$=$2$, $q$=$0.95$, $\gamma$=$0.1$. We found that with matched model parameters, SIPS achieved good performance with 10 particles per goal without the use of rejuvenation moves, so we report those results here. Further experimental details and parameters can be found in the supplement.

We summarize the results of these experiments in Table \ref{tab:performance}, with additional results in the supplement. Our method greatly outperforms the unbiased BIRL baseline in both accuracy and speed in three out of four domains, with an average runtime (AC) often several orders of magnitude smaller. This is largely because unbiased VI fails to converge except for the highly restricted Taxi domain. In contrast, SIPS requires far less initial computation, albeit with higher marginal cost due its online generation of partial plans. In fact, it achieves comparable accuracy and speed to the oracle BIRL baseline, sometimes with less computation (e.g. in Doors, Keys \& Gems). SIPS also produces higher estimates of the goal posterior $P(g_\text{true}|o)$. This is a reflection of the underlying agent model, which assumes randomness at the level of planning instead of acting. As a result, even a few observations can provide substantial evidence that a particular plan and goal was chosen.

\begin{table}[ht]
%\scriptsize
\fontsize{7pt}{7pt}\selectfont
\centering

\begin{subtable}[b]{\textwidth}
\centering
\begin{tabular}{@{}clrrrrrrrrrr@{}}
\toprule
\multicolumn{1}{l}{} &
   &
  \multicolumn{6}{c}{\textbf{Accuracy}} &
  \multicolumn{4}{c}{\textbf{Runtime}} \\ \midrule
\multirow{2}{*}{\textbf{Domain}} &
  \multicolumn{1}{c}{\multirow{2}{*}{\textbf{Method}}} &
  \multicolumn{3}{c}{$P(g_\text{true} | o)$} &
  \multicolumn{3}{c}{Top-1} &
  \multicolumn{1}{l}{\multirow{2}{*}{C$_0$ (s)}} &
  \multicolumn{1}{l}{\multirow{2}{*}{MC (s)}} &
  \multicolumn{1}{l}{\multirow{2}{*}{AC (s)}} &
  \multicolumn{1}{l}{\multirow{2}{*}{N}} \\
 &
  \multicolumn{1}{c}{} &
  \multicolumn{1}{l}{Q1} &
  \multicolumn{1}{l}{Q2} &
  \multicolumn{1}{l}{Q3} &
  \multicolumn{1}{l}{Q1} &
  \multicolumn{1}{l}{Q2} &
  \multicolumn{1}{l}{Q3} &
  \multicolumn{1}{l}{} &
  \multicolumn{1}{l}{} &
  \multicolumn{1}{l}{} &
  \multicolumn{1}{l}{} \\ \midrule \midrule
\multirow{3}{*}{\begin{tabular}[c]{@{}c@{}}Taxi\\ (3 Goals)\end{tabular}} &
  SIPS (ours) &
  \textbf{0.44} &
  \textbf{0.50} &
  0.62 &
  \textbf{0.53} &
  \textbf{0.56} &
  0.67 &
  13.0 &
  1.80 &
  2.55 &
  \textbf{1429} \\
 &
  BIRL (unbiased) &
  0.34 &
  0.35 &
  \textbf{0.79} &
  0.33 &
  0.42 &
  \textbf{0.92} &
  \textbf{2.22} &
  \textbf{0.00} &
  \textbf{0.16} &
  10000 \\
 &
  BIRL (oracle) &
  0.37 &
  0.47 &
  0.81 &
  0.42 &
  0.44 &
  0.86 &
  1.63 &
  0.00 &
  0.12 &
  2500 \\ \midrule
\multirow{3}{*}{\begin{tabular}[c]{@{}c@{}}Doors, \\ Keys \& Gems\\ (3 Goals)\end{tabular}} &
  SIPS (ours) &
  \textbf{0.37} &
  \textbf{0.51} &
  \textbf{0.61} &
  \textbf{0.74} &
  \textbf{0.74} &
  \textbf{0.74} &
  \textbf{3.30} &
  0.70 &
  \textbf{0.86} &
  \textbf{2099} \\
 &
  BIRL (unbiased) &
  0.33 &
  0.33 &
  0.33 &
  0.33 &
  0.33 &
  0.33 &
  3326 &
  \textbf{0.12} &
  154 &
  250000 \\
 &
  BIRL (oracle) &
  0.37 &
  0.36 &
  0.42 &
  0.44 &
  0.60 &
  0.80 &
  150 &
  0.12 &
  7.01 &
  10000 \\ \midrule
\multirow{3}{*}{\begin{tabular}[c]{@{}c@{}}Block Words\\ (5 Goals)\end{tabular}} &
  SIPS (ours) &
  \textbf{0.47} &
  \textbf{0.83} &
  \textbf{0.90} &
  \textbf{0.78} &
  \textbf{0.84} &
  \textbf{0.91} &
  \textbf{20.8} &
  2.46 &
  \textbf{4.15} &
  \textbf{2506} \\
 &
  BIRL (unbiased) &
  0.20 &
  0.20 &
  0.21 &
  0.42 &
  0.49 &
  0.56 &
  687 &
  \textbf{0.27} &
  63.6 &
  250000 \\
 &
  BIRL (oracle) &
  0.20 &
  0.29 &
  0.45 &
  0.73 &
  0.80 &
  0.96 &
  22.2 &
  0.05 &
  2.12 &
  10000 \\ \midrule
\multirow{3}{*}{\begin{tabular}[c]{@{}c@{}}Intrusion \\ Detection\\ (20 Goals)\end{tabular}} &
  SIPS (ours) &
  \textbf{0.56} &
  \textbf{0.87} &
  \textbf{0.87} &
  \textbf{0.65} &
  \textbf{0.87} &
  \textbf{0.87} &
  \textbf{375} &
  6.60 &
  \textbf{28.0} &
  \textbf{13321} \\
 &
  BIRL (unbiased) &
  0.05 &
  0.05 &
  0.05 &
  0.05 &
  0.05 &
  0.05 &
  18038 &
  \textbf{0.75} &
  1069 &
  250000 \\
 &
  BIRL (oracle) &
  0.09 &
  0.24 &
  0.53 &
  0.94 &
  1.00 &
  1.00 &
  98 &
  0.02 &
  6.00 &
  10000 \\ \bottomrule
\end{tabular}
\caption{\textbf{Accuracy and runtime of goal inference across domains and inference methods.} We quantify accuracy at the 1st, 2nd and 3rd quartiles (Q1--Q3) of each observed trajectory via the posterior probability of the true goal $P(g_\text{true}|o)$, and the fraction of problems where $g_\text{true}$ is top-ranked (Top-1). We measure runtime in terms of the start-up cost (C$_0$), marginal cost per timestep (MC), and average cost per timestep (AC) in seconds. We also report the total number of states visited (N) during either search or value iteration as a platform-independent measure. Excluding the oracle baseline, the best metrics are bolded.}
\label{tab:performance}
\end{subtable}

\begin{subtable}[b]{\textwidth}
\centering
%\scriptsize
\begin{tabular}{@{}lrrrrrrrl@{}}
\toprule
 &
  \multicolumn{3}{c}{\textbf{Persistence ($r$)}} &
  \multicolumn{3}{c}{\textbf{Persistence ($q$)}} &
  \multicolumn{1}{c}{\textbf{RL}} &
  \multicolumn{1}{c}{\textbf{Optimal}} \\
\textbf{Domain}   & 1    & 2*   & 4    & 0.8  & 0.9  & 0.95* & $\alpha$=50     &       \\ \midrule
Doors, Keys, Gems & 0.60 & 0.73 & 0.73 & 0.53 & 0.60 & 0.73  & 0.58            & 0.80  \\
Block Words       & 0.90 & 0.87 & 0.90 & 0.70 & 0.83 & 0.87  & 0.82            & 0.80  \\ \midrule
 &
  \multicolumn{3}{c}{\textbf{Search Noise. ($\gamma$)}} &
  \multicolumn{4}{c}{\textbf{Heuristic ($h$)}} &
  \multicolumn{1}{c}{\textbf{Humans}} \\
\textbf{Domain}   & 0.5  & 0.1* & 0.02 & Mh.* & Mz.  & GC.   & $h_\text{add}$* & $n$=5 \\ \midrule
Doors, Keys, Gems & 0.67 & 0.73 & 0.77 & 0.73 & 0.90 & --    & --              & 0.79  \\
Block Words       & 0.83 & 0.87 & 0.87 & --   & --   & 0.43  & 0.87            & 0.73  \\ \bottomrule
\end{tabular}
\caption{{\bf Robustness to model mismatch.} Top-1 accuracy of SIPS at the third time quartile (Q3), evaluated on data generated by mismatched parameters, Boltzmann-rational RL agents, optimal agents, and humans. We ran SIPS assuming $r$=$2$, $q$=$0.95$, $T$=$10$. For Doors, Keys, Gems, we assumed a Manhattan (Mh.) heuristic against a maze distance (Mz.) heuristic. For Block Words, we assumed $h_\text{add}$ against the naive goal count (GC.) heuristic. Matched parameters are starred (*).}
\label{tab:robustness}
\end{subtable}
\caption{Accuracy, runtime, and robustness of inference.}
\end{table}

Given the specific assumptions made by our agent model, a reasonable question is whether inference is robust to plans generated by other agent models or actual humans. To address this, we also performed a series of robustness experiments for two domains (Table \ref{tab:robustness}) on data generated by mismatched model parameters $r$, $q$, $\gamma$, mismatched planning heuristics $h$, Boltzmann-rational RL agents, optimal agents, and 5 pilot human subjects (30 trajectories per subject).

As Table \ref{tab:robustness} shows, SIPS is relatively robust to data generated by these other models and parameters. Although performance can degrade with mismatch, this is partly due to the difficulty of inference from highly random behavior (e.g. $q$=$0.8$, $h$=GC.). On the other hand, when mismatched parameters are \emph{more} optimal, performance can \emph{improve} (e.g. $h$=Mz.). Importantly, SIPS also does well on human data, showing robustness even when the planner is unknown. While our boundedly rational agent model cannot possibly capture all aspects of human planning, these experiments suggest that it is serves as a reasonable approximation, similar to our intuitive theories of other people's minds.

\section{Limitations and Future Work}

In this paper, we demonstrated an architecture capable of online inference of  goals and plans, even when those plans might fail. However, important limitations remain. First, we considered only finite sets of goals, but the space of goals that humans pursue is easily infinite. Relatedly, we assume that these goals are final, instead of accounting for the hierarchical and instrumental nature of goals and plans. A promising next step would thus be to express hierarchies of goals and plans as probabilistic grammars or programs \cite{lake2015human,saad2019bayesian,ellis2020dreamcoder}, capturing both the infinitude and structure of the motives we attribute to each other \cite{cundy2018exploring,kaelbling2011hierarchical}. Second, unlike the domains considered here, the environments we operate in often involve stochastic dynamics and infinite action spaces \cite{garrett2017strips,garrett2020pddlstream}. A natural extension would be to integrate Monte Carlo Tree Search or sample-based motion planners into our architecture as modeling components \cite{cusumano2017probabilistic}, potentially parameterized by learned heuristics \cite{silver2016mastering}. With hope, our architecture might then approach the full complexity of problems that we face everyday, whether one is stacking blocks as a kid, finding the right keys for the right doors, or writing a research paper.

\pagebreak

\section{Broader Impact}

We embarked upon this research in the belief that, as increasingly powerful autonomous systems become embedded in our society, it may eventually become necessary for them to accurately understand our goals and values, so as to robustly act in our collective interest. Crucially, this will require such systems to understand the ways in which humans routinely fail to achieve our goals, and not take that as evidence that those goals were never desired. Due to our manifold cognitive limitations, gaps emerge between our goals and our intentions, our intentions and our actions, our beliefs and our conclusions, and our ideals and our practices. To the extent that we would like machines to aid us in actualizing the goals and ideals we most value, rather than those we appear to be acting towards, it will be critical for them to understand how, when, and why those gaps emerge. This aspect of the value alignment problem has thus far been under-explored \cite{russell2019human}. By performing this research at the intersection of cognitive science and AI, we hope to lay some of the conceptual and technical groundwork that may be necessary to understand our boundedly-rational behavior.

Of course, the ability to infer the goals of others, and to do so online and despite failures, has many more immediate uses, each of them with its own set of benefits and risks. Perhaps the most straightforwardly beneficial are assistive use cases, such as smart user interfaces \cite{horvitz2013lumiere}, intelligent personal assistants, and collaborative robots, which may offer to aid a user if that user appears to be pursuing a sub-optimal plan. However, even those use cases come with the risk of reducing human autonomy, and care should be taken so that such applications ensure the autonomy and willing consent of those being aided \cite{sarathy2019exceptions}.

More concerning however is the potential for such technology to be abused for manipulative, offensive, or surveillance purposes. While the research presented in this paper is nowhere near the level of integration that would be necessary for active surveillance or manipulation, it is highly likely that mature versions of similar technology will be co-opted for such purposes by governments, militaries, and the security industry \cite{zuboff2015big,brundage2018malicious}. Although detecting and inferring ``suspicious intent'' may not seem harmful in its own right, these uses need to be considered within the broader context of society, especially the ways in which marginalized peoples are over-policed and incarcerated \cite{ferguson2019rise}. Given these risks, we urge future research on this topic to consider seriously the ways in which technology of this sort will most likely be used, by which institutions, and whether those uses will tend to lead to just and beneficial outcomes for society as a whole. The ability to infer and understand the motives of others is a skill that can be wielded to both great benefit and great harm. We ought to use it wisely.

\section{Code Availability}

Code for the architecture and experiments presented in this paper is available at \url{https://github.com/ztangent/Plinf.jl/tree/neurips-2020-experiments}, as part of the \texttt{Plinf.jl} package for Bayesian inverse planning.

\section{Acknowledgements}

This work was funded in part by the DARPA Machine Common Sense program (Award ID: 030523-00001); philanthropic gifts from the Aphorism Foundation and from the Siegel Family Foundation; and financial support from the MIT-IBM Watson AI Lab and the Intel Probabilistic Computing Center. Tom Silver is supported by an NSF Graduate Research Fellowship.

\end{document}

\maketitle

\appendix
\makeatletter
\renewcommand{\thefigure}{S\arabic{figure}}
\renewcommand{\thetable}{S\arabic{table}}
\makeatletter

\section{Experimental Details}

Below we provide experimental details for each of the inference methods described in the main text. We have also performed additional experiments using a baseline adapted from the plan recognition as planning (PRP) literature \cite{ramirez2010probabilistic}, which we include below as a useful offline benchmark.

\subsection{Sequential Inverse Plan Search}

We conducted experiments using two main variants of Sequential Inverse Plan Search (SIPS), the first using data-driven rejuvenation, as described in the main text, and the second without. Rejuvenation is necessary for the results shown in Figure 1 of the main text, and for highly sub-optimal and failed plans more generally. However, rejuvenation is also hard to tune, and can increase runtime due to the need to replan. We thus report results without rejuvenation in our quantitative experiments.

Parameters for qualitative experiments are given in each of the corresponding figures in section B.1. For the quantitative experiments, we used SIPS with 10 particles per possible goal (e.g., 50 particles for the Block Words domain), with a resampling threshold of $c=1/4$. For the underlying agent model, we assumed search noise of $\gamma=0.1$ and persistence parameters of $r=2$ and $q=0.95$ (giving an average search budget of $38$ nodes). We varied the search heuristic $h$ to suit the type of domain: For the gridworld-based domains (Taxi; Doors, Keys \& Gems), we used a Manhattan distance heuristic to the goal. For the other domains (Block Words; Intrusion Detection), we used the $h_\text{add}$ heuristic introduced by the HSP algorithm \cite{bonet2001planning} as a generalized relaxed-distance heuristic.

SIPS also requires the specification of an observation model $P(o|s)$, in order to score the likelihood of a hypothesized state trajectory $\hat s_1, ..., \hat s_t$ given the observed states $o_1, ..., o_t$. We defined this observation model by adding zero-mean Gaussian noise with $\sigma=0.25$ for each numeric variable in the state (e.g., the agent's position in a gridworld), and Bernoulli corruption noise with $p=0.05$ for each Boolean variable in the state (e.g. whether block A is on top of block B).

All SIPS experiments were performed using \texttt{Plinf.jl}, a Julia implementation of our modeling and inference architecture that integrates the Gen probabilistic programming system with \texttt{PDDL.jl} , a Julia interpreter for the Planning Domain Definition Language \cite{mcdermott1998pddl}. Experiments were run on a 1.9 GHz Intel Core i7 processor with 16 GB RAM.

\subsection{Bayesian Inverse Reinforcement Learning}

Bayesian Inverse Reinforcement Learning (BIRL) requires computing an approximate value function $Q(s, a)$ offline and a posterior over goals online using the likelihood $P(a \mid s, g) = \frac{1}{Z}e^{\alpha \cdot Q(s, a)}$, where $Z$ is the partition function and $\alpha$ is an optimality parameter. For the quantitative experiments, we used $\alpha=1$ which we found to perform well in preliminary trials. For qualitative comparisons, however, we used $\alpha = 5$, as this choice produced results in range more similar to human inferences. To approximate the value function, we used value iteration (VI) with a discount factor of $0.9$.

As discussed in the main text, several of the domains considered in this work have state spaces that are too large to enumerate, making standard VI intractable. We therefore used asynchronous VI, sampling states instead of fully enumerating them, for 250,000 iterations for the unbiased baseline (BIRL-U). Preliminary experiments suggested that running for up to 1,000,000 iterations did not appreciably improve results. Taxi, which has a far smaller state space than the other domains, was run with 10,000 iterations, which was consistently sufficient for convergence.
For the oracle baseline (BIRL-O), 2500 iterations were sufficient to reach convergence for the Taxi domain, and 10,000 iterations for the other domains.

All BIRL experiments were written in Python and run on a 2.9 GHz Intel Core i9 processor with 32 GB RAM. We made use of the PDDLGym library \cite{silver2020pddlgym} for instantiating the PDDL planning problems as OpenAI Gym environments.
To perform asynchronous VI efficiently, we implemented state samplers and valid action generators for each domain. The unbiased version of BIRL (BIRL-U) uses these state samplers to sample states within asynchronous VI. For the oracle baseline (BIRL-O), which has access to the test-time trajectories, we instead sampled one state uniformly at random from the states visited across all test-time trajectories.

\subsection{Plan Recognition as Planning}

We adapted the plan recognition as planning (PRP) approach described in \cite{ramirez2010probabilistic} as an offline benchmark that achieves high accuracy at the cost of substantially more runtime (up to 30 times) than SIPS. In the PRP approach, we use a heuristic approximation to the likelihood of a plan $p$ given a goal $g$:
\begin{equation}
    P(p|g) \propto e^{-\beta (|p| - |p^g_*|)}
\end{equation}
where $p^g_*$ is an optimal plan to the goal $g$, $|p|$ denotes the length of the plan $p$, and $\beta$ is a noise parameter. This likelihood function model agent rationality by placing exponentially less probability on costlier plans, where larger values of $\beta$ correspond to more optimality.

In order to perform inference using this likelihood model, we first compute the optimal plan $p^g_*$ for each possible goal $g$ in a domain. At each timestep $t$, we then construct a plan $p^g_t$ to each goal $g$ consistent with the observations so far, by computing an optimal partial plan $p_t^+$ from the current observed state $o_t$ to $g$, and then concatenating it with the initial sequence of actions $p_t^- := a_1, ..., a_{t-1}$ taken by the agent, giving $p_t^g = [p_t^-, p_t^+]$. Under the additional approximation that $p_t^g$ is the only plan consistent with the observation sequence $o_1, ..., o_t$, we can then compute the goal posterior as
\begin{equation}
    P(g|o_1, ..., o_t) \simeq \frac{e^{-\beta (|p^g_t| - |p^g_*|)}}{\sum_{g' \in \mathcal{G}} e^{-\beta (|p^{g'}_t| - |p^{g'}_*|)}}
\end{equation}

The main limitation of this approach is that it requires computation of an optimal partial plan $p_t^+$ for every goal $g$ at every timestep $t$, which scales poorly with the number of goals and timesteps per trajectory, especially when the observed trajectory leads the agent further and further away from most of the goals under consideration. This is contrast to SIPS, which performs incremental computation by extending partial plans from previous timesteps. In addition, due to the assumption that there always exists a plan from the current observed state $o_t$ to every goal $g$, the PRP approach is unable to account for irreversible failures. This is shown in our qualitative comparisons.

Nonetheless, because PRP still achieves high accuracy on many sub-optimal trajectories (at the expense of considerably more computation, especially on domains with many goals), we include it here as a benchmark for accuracy. All PRP experiments were performed on the same machine as the SIPS experiments, using the implementation of A* search provided by \texttt{Plinf.jl}.

\section{Additional Results}

\subsection{Qualitative Comparisons for Sub-Optimal \& Failed Plans}

\begin{figure}[H]
    \centering
    \includegraphics[width=0.95\textwidth]{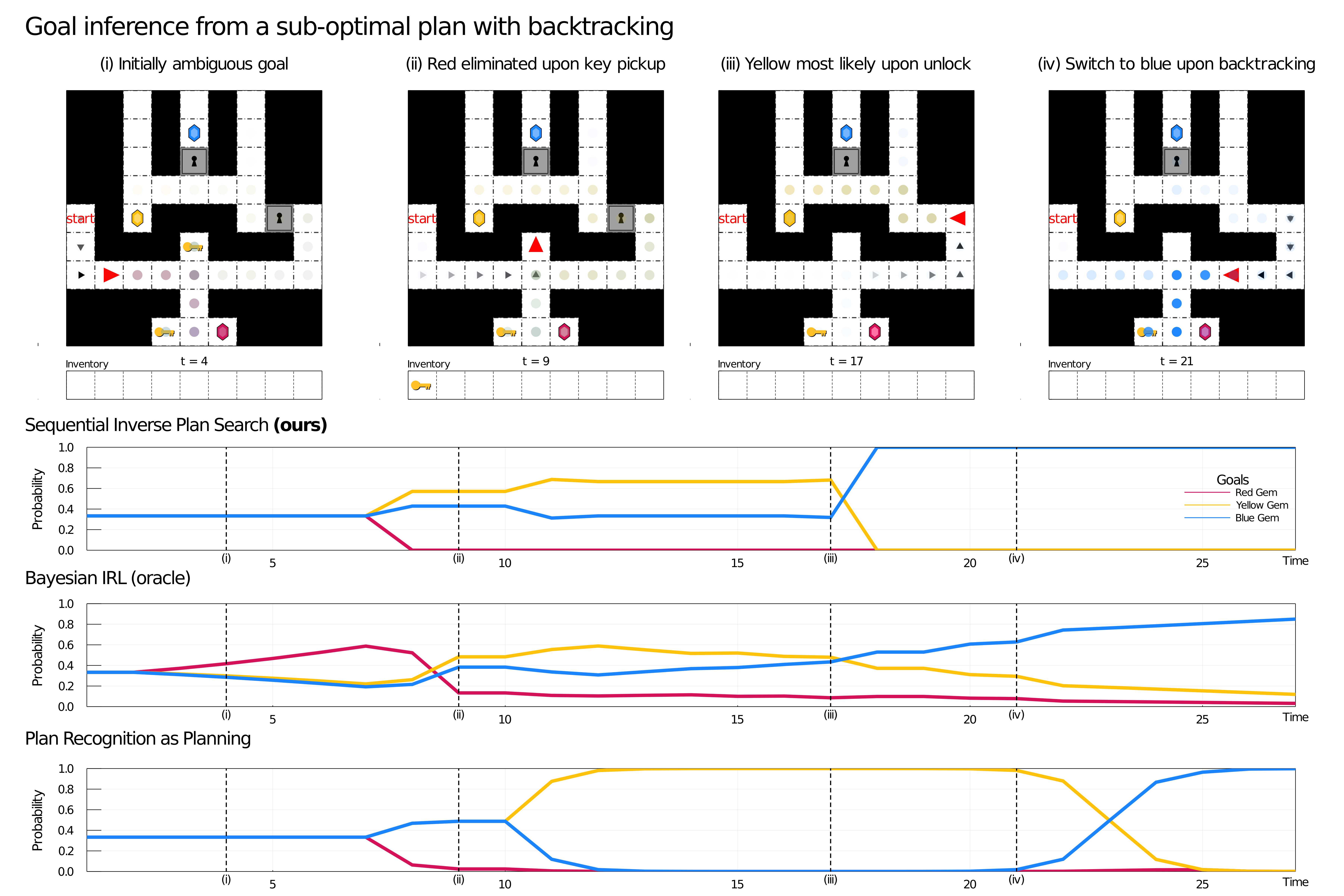}
    \caption{Goal inferences made by SIPS, BIRL-O, and PRP for the sub-optimal trajectory shown in Figure 1(a) of the main text. Predicted future trajectories in panels (i)--(iv) are made by SIPS. For SIPS, we used 30 particles per goal, search noise $\gamma=0.1$, persistence parameters $r=2$, $q=0.95$, and a Manhattan distance heuristic to the goal. Rejuvenation moves were used, with a goal rejuvenation probability of $p_g = 0.25$. For BIRL-O, we used $\alpha=5$. For PRP, we used $\beta = 1$.}
    \label{fig:comparison-1}
\end{figure}

\begin{figure}[H]
    \centering
    \includegraphics[width=0.95\textwidth]{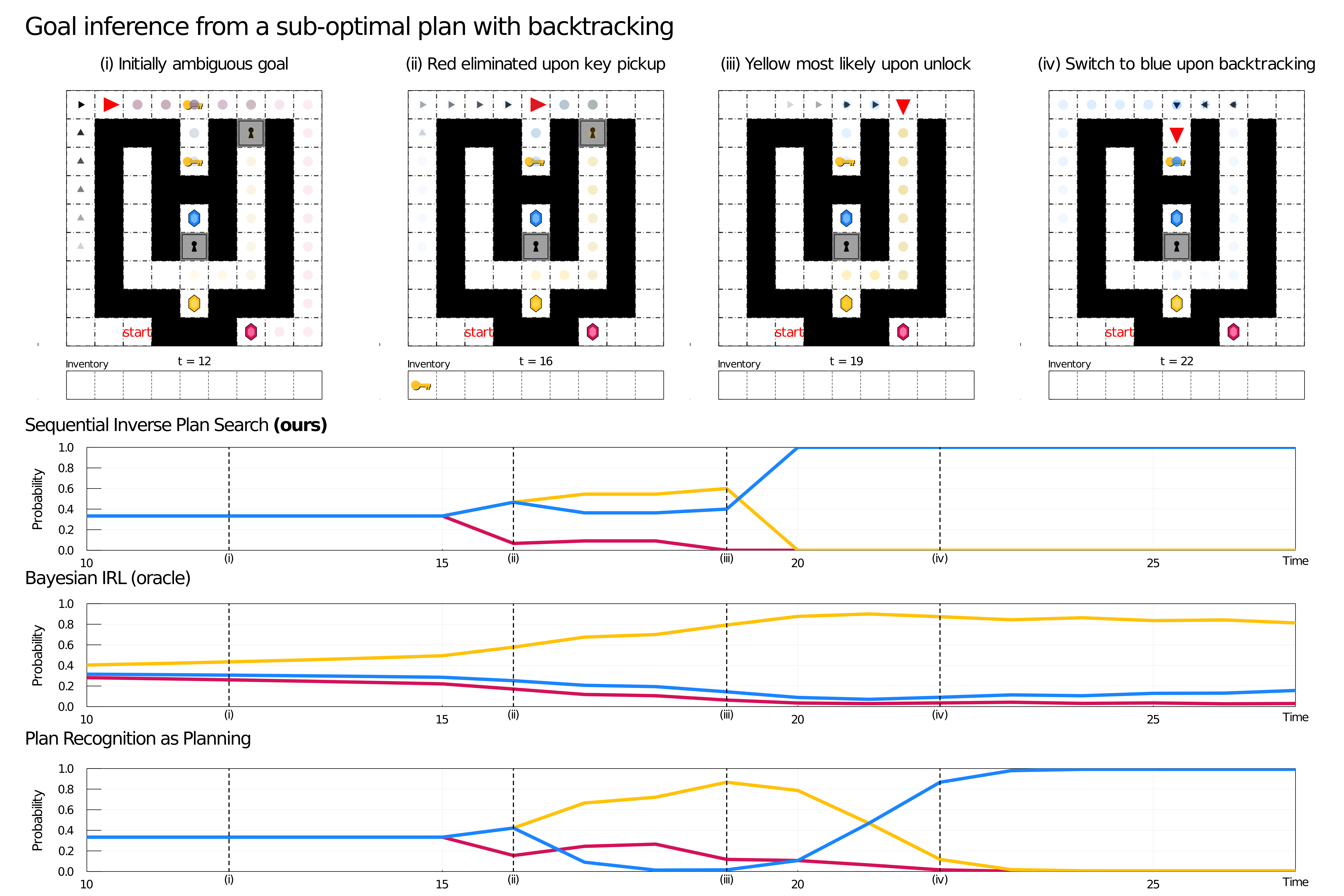}
    \caption{Goal inferences made by SIPS, BIRL-O, and PRP for another sub-optimal trajectory. Predicted future trajectories in panels (i)--(iv) are made by SIPS. For SIPS, we used 30 particles per goal, search noise $\gamma=0.1$, persistence parameters $r=2$, $q=0.95$, and a Manhattan distance heuristic to the goal. Rejuvenation moves were used, with a goal rejuvenation probability of $p_g = 0.25$. For BIRL-O, we used $\alpha=5$. For PRP, we used $\beta = 1$.}
    \label{fig:comparison-2}
\end{figure}

Here we present detailed qualitative comparisons of the goal inferences made for sub-optimal and failed plans in the Doors, Keys \& Gems domain. Figures \ref{fig:comparison-1} and \ref{fig:comparison-2} show the inferences made for two sub-optimal trajectories, while Figures \ref{fig:comparison-3} and \ref{fig:comparison-4} show the inferences made for two trajectories with irreversible failures. We omit the unbiased Bayesian IRL baseline (BIRL-U), because it is unable to solve the underlying Markov Decision Process in any of these examples, leading to a uniform posterior over goals over the entire trajectory.

\pagebreak

\subsubsection{Sub-Optimal Plans}

Figure \ref{fig:comparison-1} shows how the inferences produced by SIPS are more human-like, compared to the BIRL and PRP baselines. In particular, SIPS adjusts its inferences in a human-like manner, initially remaining uncertain between the 3 gems (panel i), placing more posterior mass on the yellow gem when the agent acquires the first key (panel ii), increasing that posterior mass when agent appears to ignore the second key and unlock the first door (panel iii), but then switching to the blue gem once the agent backtracks towards the second key (panel iv).

While the inferences produced by BIRL display similar trends, they are much more gradual, because BIRL assumes noise at the level of acting instead of planning. In addition, the agent model underlying BIRL leads to strange artifacts, such as the rise in probability of the red gem when $t<9$. This is because Boltzmann action noise places lower probability $P(a|g)$ on an action $a$ that leads to a goal $g$ which is further away, due to the value function $V_g$ associated with that goal $g$ being smaller due to time discounting. As a result, when $t < 9$, BIRL computes that $P(\texttt{right}|\texttt{red}) > P(\texttt{right}|\texttt{yellow})$ and  $P(\texttt{right}|\texttt{blue})$, leading to the red gem being inferred as the most likely goal.

Finally, PRP exhibits both over-confidence in the yellow gem and slow recovery towards the blue gem. This is due to the assumption that the likelihood of a plan $p$ to some goal $g$ is exponentially decreasing in its cost difference from the optimal plan $p^g_*$. Between $t=10$ and $t=20$, all plans consistent with the observations to the blue gem are considerably longer than the optimal plan $p^\texttt{blue}_*$. As a result, PRP gives very low probability to the blue gem. This effect continues for many timesteps after the agent starts to backtrack ($t=17$ to $t=24$), indicating that the PRP modeling assumptions are inadequate for plans with substantial backtracking.

Similar dynamics can be observed for the trajectory in Figure \ref{fig:comparison-2}. The BIRL baseline performs especially poorly, placing high probability on the yellow gem even when the agent backtracks to collect the second key ($t=19$ to $t=22$). This again is due to the assumption of action noise instead of planning noise, making it much more likely under the BIRL model that an agent would randomly walk back towards the second key. The PRP baseline exhibits the same issues with over-confidence and slow recovery described earlier, placing so little posterior mass on the blue gem from $t = 17$ to $t= 20$ that it even considers the red gem to be more likely. In contrast, our method, SIPS, immediately converges to the blue gem once backtracking occurs at $t=20$.

\subsubsection{Failed Plans}

\begin{figure}[H]
    \centering
    \includegraphics[width=0.95\textwidth]{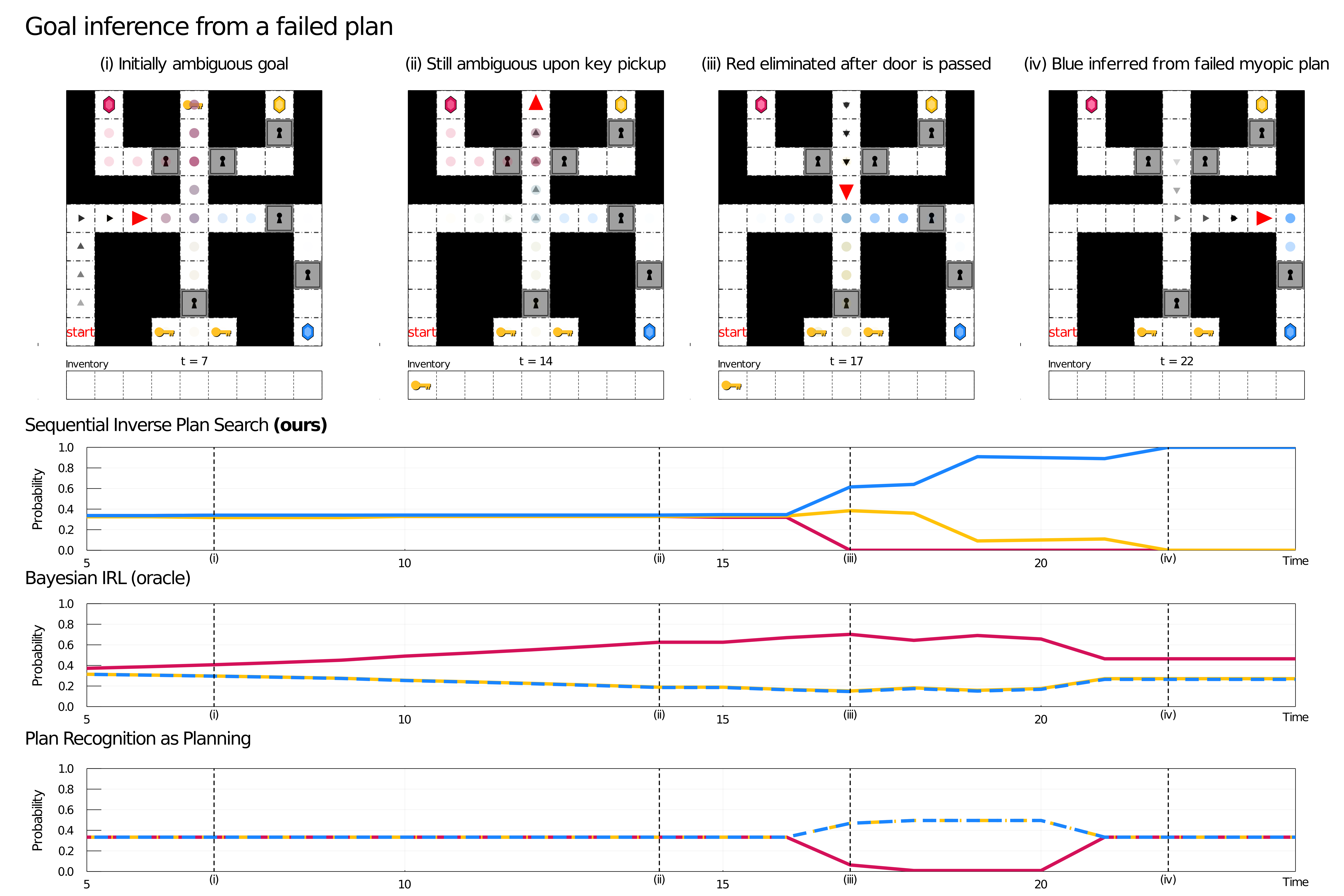}
    \caption{Goal inferences made by SIPS, BIRL-O, and PRP for the failed trajectory shown in Figure 1(b) of the main text. Predicted future trajectories in panels (i)--(iv) are made by SIPS. For SIPS, we used 30 particles per goal, search noise $\gamma=0.1$, persistence parameters $r=2$, $q=0.95$, and a maze-distance heuristic (i.e. distance to the goal, ignoring doors). Rejuvenation moves were used with $p_g = 0.25$. For BIRL-O, we used $\alpha=5$. For PRP, we used $\beta = 1$.}
    \label{fig:comparison-3}
\end{figure}

The differences between SIPS and the baseline methods are even more striking for trajectories with irreversible failures. As shown in Figure \ref{fig:comparison-3}, SIPS accurately infers that the blue gem is the most likely goal when the agent ignores the two keys at the bottom, instead turning towards the first door guarding the blue gem at $t=19$. This inference also remains stable after $t=21$, when the agent irreversibly uses up its key to unlock that door. SIPS is capable of such inferences because the search for partial plans is biased towards promising intermediate states. Since the underlying agent model assumes a relaxed distance heuristic that considers states closer to the blue gem as promising, the model is likely to produce partial plans that lead spatially toward the blue gem, even if those plans myopically use up the agent's only key.

In contrast, both BIRL and PRP fail to infer that the blue gem is the goal. BIRL initially places increasing probability on the red gem, due to Boltzmann action noise favoring goals which take less time to reach. While this probability decreases slightly as the agent detours from the optimal plan to the red gem, it remains the highest probability goal even after the agent uses up its key at $t=21$. The posterior over goals stops changing after that, because there are no longer any any possible paths to a goal. PRP exhibits a different failure mode. While it does not suffer from the artifacts due to Boltzmann action noise, it completely fails to account for the possibility that an agent might make a failed plan. As a result, the probability of the blue gem does not increase even after the agent turns towards it at $t=19$. Furthermore, once failure occurs at $t=21$, PRP ends up defaulting to a uniform distribution over the three gems, even though it had previously eliminated the red gem as a possibility.

\begin{figure}[H]
    \centering
    \includegraphics[width=0.95\textwidth]{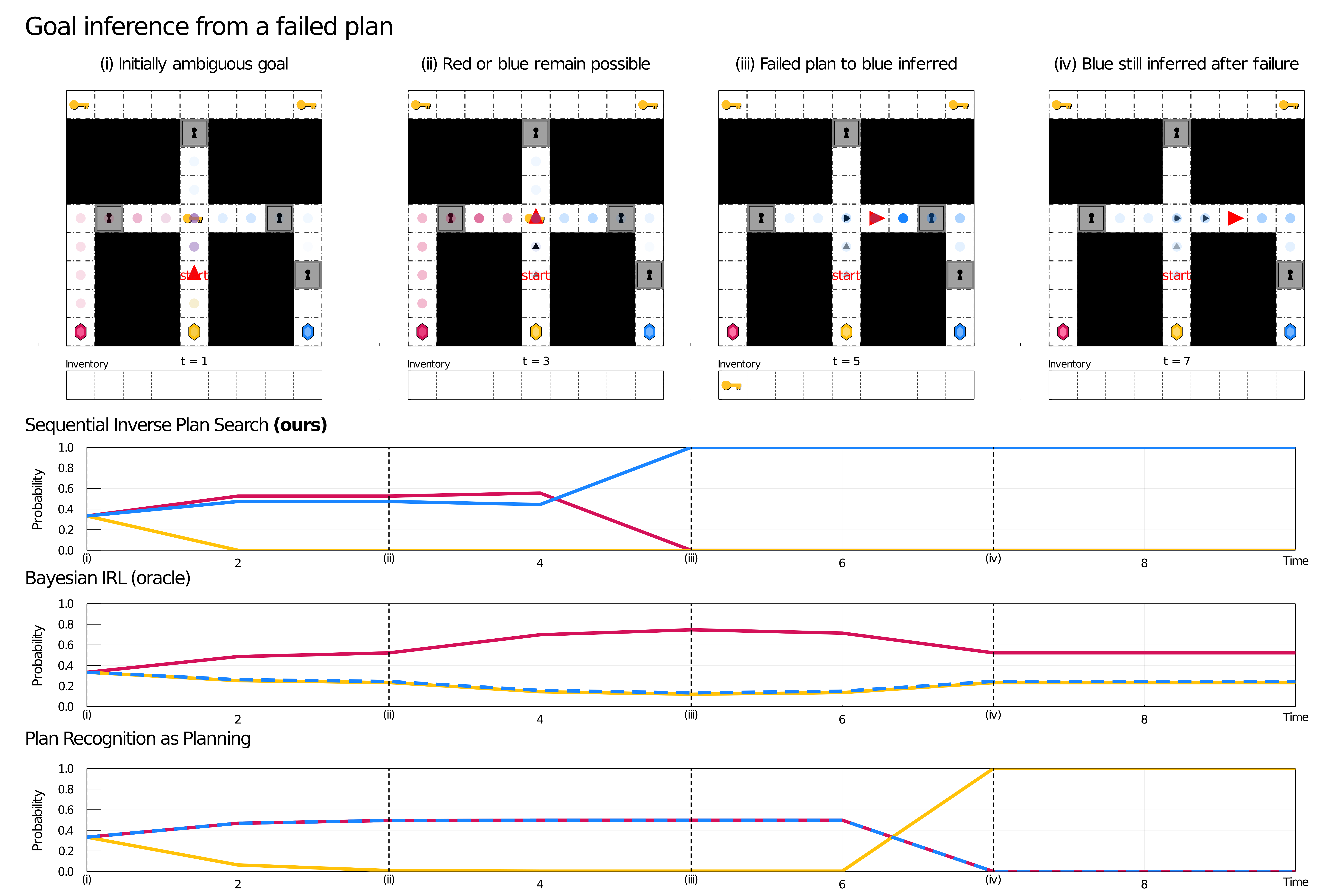}
    \caption{Goal inferences made by SIPS, BIRL-O, and PRP for another failed trajectory. Predicted future trajectories in panels (i)--(iv) are made by SIPS. For SIPS, we used 30 particles per goal, search noise $\gamma=0.1$, persistence parameters $r=2$, $q=0.95$, and a Manhattan distance heuristic to the goal. Rejuvenation moves were used, with a goal rejuvenation probability of $p_g = 0.25$. For BIRL-O, we used $\alpha=5$. For PRP, we used $\beta = 1$.}
    \label{fig:comparison-4}
\end{figure}

The inferences in Figure \ref{fig:comparison-4}
display similar trends. Once again, SIPS accurately infers that the blue gem is the goal, even slightly in advance of failure (panel iii). In contrast, BIRL wrongly infers that the red gem is the most likely, while PRP erroneously defaults to inferring upon failure that the only remaining acquirable gem (yellow) is the goal.

\pagebreak

\pagebreak

\subsection{Accuracy \& Speed}
Here we present quantitative comparisons of the accuracy and speed of each inference method. Tables \ref{tab:accuracy-optimal-dataset} and \ref{tab:accuracy-suboptimal-dataset} show the accuracy results for the optimal and sub-optimal datasets respectively. $P(g_\text{true}|o)$ represents the posterior probability of the true goal, while Top-1 represents the fraction of problems where $g_\text{true}$ is top-ranked. Accuracy metrics are reported at the first (Q1), second (Q2), and third (Q3) quartiles of each observed trajectory. The corresponding standard deviations (taken across the dataset) are shown to the right of each accuracy mean.

Tables \ref{tab:runtime-optimal-dataset} and \ref{tab:runtime-suboptimal-dataset} show the runtime results for the optimal and sub-optimal datasets respectively. Runtime is reported in terms of the start-up cost (C$_0$), marginal cost per timestep (MC), and average cost per timestep (AC), all measured in seconds. The corresponding standard deviations are shown to the right of each runtime mean. The total number (N) of states visited (during either plan search or value iteration) are also reported as a platform-independent cost metric.

\begin{table}[h]
\centering
\scriptsize
\begin{tabular}{@{}clrrrrrrrrrrrr@{}}
\toprule
\multicolumn{1}{l}{} &
   &
  \multicolumn{12}{c}{\textbf{Accuracy}} \\ \midrule
\textbf{Domain} &
  \multicolumn{1}{c}{\textbf{Method}} &
  \multicolumn{6}{c}{$P(g_\text{true}| o)$} &
  \multicolumn{6}{c}{Top-1} \\
\multicolumn{1}{l}{\textbf{}} &
  \textbf{} &
  \multicolumn{2}{c}{Q1} &
  \multicolumn{2}{c}{Q2} &
  \multicolumn{2}{c}{Q3} &
  \multicolumn{2}{c}{Q1} &
  \multicolumn{2}{c}{Q2} &
  \multicolumn{2}{c}{Q3} \\ \midrule
\multirow{4}{*}{\begin{tabular}[c]{@{}c@{}}Taxi\\ (3 Goals)\end{tabular}} &
  SIPS &
  0.45 &
  $\pm$0.26 &
  0.48 &
  $\pm$0.27 &
  0.64 &
  $\pm$0.32 &
  0.67 &
  $\pm$0.49 &
  0.67 &
  $\pm$0.49 &
  0.67 &
  $\pm$0.49 \\
 &
  BIRL-U &
  0.33 &
  $\pm$0.06 &
  0.38 &
  $\pm$0.17 &
  0.79 &
  $\pm$0.22 &
  0.33 &
  $\pm$0.47 &
  0.42 &
  $\pm$0.49 &
  0.92 &
  $\pm$0.28 \\
 &
  BIRL-O &
  0.41 &
  $\pm$0.33 &
  0.44 &
  $\pm$0.40 &
  0.82 &
  $\pm$0.23 &
  0.50 &
  $\pm$0.50 &
  0.42 &
  $\pm$0.49 &
  1.00 &
  $\pm$0.00 \\
 &
  PRP &
  0.33 &
  $\pm$0.00 &
  0.36 &
  $\pm$0.06 &
  0.44 &
  $\pm$0.08 &
  0.33 &
  $\pm$0.00 &
  1.00 &
  $\pm$0.00 &
  1.00 &
  $\pm$0.00 \\ \midrule
\multirow{4}{*}{\begin{tabular}[c]{@{}c@{}}Doors,\\ Keys \& \\ Gems\\ (3 Goals)\end{tabular}} &
  SIPS &
  0.39 &
  $\pm$0.18 &
  0.51 &
  $\pm$0.32 &
  0.70 &
  $\pm$0.35 &
  0.73 &
  $\pm$0.46 &
  0.73 &
  $\pm$0.46 &
  0.80 &
  $\pm$0.41 \\
 &
  BIRL-U &
  0.33 &
  $\pm$0.00 &
  0.33 &
  $\pm$0.00 &
  0.33 &
  $\pm$0.00 &
  0.33 &
  $\pm$0.00 &
  0.33 &
  $\pm$0.00 &
  0.33 &
  $\pm$0.00 \\
 &
  BIRL-O &
  0.41 &
  $\pm$0.33 &
  0.37 &
  $\pm$0.06 &
  0.41 &
  $\pm$0.08 &
  0.50 &
  $\pm$0.50 &
  0.67 &
  $\pm$0.47 &
  0.87 &
  $\pm$0.34 \\
 &
  PRP &
  0.40 &
  $\pm$0.17 &
  0.62 &
  $\pm$0.30 &
  0.81 &
  $\pm$0.26 &
  1.00 &
  $\pm$0.00 &
  1.00 &
  $\pm$0.00 &
  1.00 &
  $\pm$0.00 \\ \midrule
\multirow{4}{*}{\begin{tabular}[c]{@{}c@{}}Block\\ Words\\ (5 Goals)\end{tabular}} &
  SIPS &
  0.38 &
  $\pm$0.27 &
  0.71 &
  $\pm$0.41 &
  0.78 &
  $\pm$0.41 &
  0.73 &
  $\pm$0.46 &
  0.73 &
  $\pm$0.46 &
  0.80 &
  $\pm$0.41 \\
 &
  BIRL-U &
  0.20 &
  $\pm$0.03 &
  0.21 &
  $\pm$0.05 &
  0.23 &
  $\pm$0.10 &
  0.53 &
  $\pm$0.50 &
  0.53 &
  $\pm$0.50 &
  0.60 &
  $\pm$0.49 \\
 &
  BIRL-O &
  0.22 &
  $\pm$0.01 &
  0.30 &
  $\pm$0.03 &
  0.46 &
  $\pm$0.06 &
  0.73 &
  $\pm$0.44 &
  0.87 &
  $\pm$0.34 &
  1.00 &
  $\pm$0.00 \\
 &
  PRP &
  0.38 &
  $\pm$0.18 &
  0.78 &
  $\pm$0.28 &
  0.91 &
  $\pm$0.18 &
  0.93 &
  $\pm$0.26 &
  0.93 &
  $\pm$0.26 &
  1.00 &
  $\pm$0.00 \\ \midrule
\multirow{4}{*}{\begin{tabular}[c]{@{}c@{}}Intrusion\\ Detection\\ (20 Goals)\end{tabular}} &
  SIPS &
  0.65 &
  $\pm$0.38 &
  1.00 &
  $\pm$0.00 &
  1.00 &
  $\pm$0.00 &
  0.80 &
  $\pm$0.41 &
  1.00 &
  $\pm$0.00 &
  1.00 &
  $\pm$0.00 \\
 &
  BIRL-U &
  0.05 &
  $\pm$0.00 &
  0.05 &
  $\pm$0.00 &
  0.05 &
  $\pm$0.00 &
  0.05 &
  $\pm$0.00 &
  0.05 &
  $\pm$0.00 &
  0.05 &
  $\pm$0.00 \\
 &
  BIRL-O &
  0.10 &
  $\pm$0.01 &
  0.25 &
  $\pm$0.02 &
  0.55 &
  $\pm$0.03 &
  1.00 &
  $\pm$0.00 &
  1.00 &
  $\pm$0.00 &
  1.00 &
  $\pm$0.00 \\
 &
  PRP &
  0.35 &
  $\pm$0.13 &
  0.96 &
  $\pm$0.06 &
  0.99 &
  $\pm$0.01 &
  1.00 &
  $\pm$0.00 &
  1.00 &
  $\pm$0.00 &
  1.00 &
  $\pm$0.00 \\ \bottomrule
\end{tabular}
\vspace{6pt}
\caption{Inference accuracy on the dataset of optimal trajectories.}
\label{tab:accuracy-optimal-dataset}
\end{table}

\begin{table}[h]
    \centering
    \scriptsize
    \begin{tabular}{@{}clllllllllllll@{}}
    \toprule
    \multicolumn{1}{l}{} &
       &
      \multicolumn{12}{c}{\textbf{Accuracy}} \\ \midrule
    \textbf{Domain} &
      \multicolumn{1}{c}{\textbf{Method}} &
      \multicolumn{6}{c}{$P(g_\text{true}| o)$} &
      \multicolumn{6}{c}{Top-1} \\
    \multicolumn{1}{l}{} &
       &
      \multicolumn{2}{c}{Q1} &
      \multicolumn{2}{c}{Q2} &
      \multicolumn{2}{c}{Q3} &
      \multicolumn{2}{c}{Q1} &
      \multicolumn{2}{c}{Q2} &
      \multicolumn{2}{c}{Q3} \\ \midrule
    \multirow{4}{*}{\begin{tabular}[c]{@{}c@{}}Taxi\\ (3 Goals)\end{tabular}} &
      SIPS &
      0.43 &
      $\pm$0.32 &
      0.51 &
      $\pm$0.38 &
      0.62 &
      $\pm$0.42 &
      0.46 &
      $\pm$0.51 &
      0.50 &
      $\pm$0.51 &
      0.67 &
      $\pm$0.48 \\
     &
      BIRL-U &
      0.34 &
      $\pm$0.06 &
      0.33 &
      $\pm$0.00 &
      0.79 &
      $\pm$0.23 &
      0.33 &
      $\pm$0.47 &
      0.42 &
      $\pm$0.49 &
      0.92 &
      $\pm$0.28 \\
     &
      BIRL-O &
      0.35 &
      $\pm$0.29 &
      0.48 &
      $\pm$0.32 &
      0.81 &
      $\pm$0.32 &
      0.38 &
      $\pm$0.48 &
      0.46 &
      $\pm$0.50 &
      0.79 &
      $\pm$0.41 \\
     &
      PRP &
      0.33 &
      $\pm$0.00 &
      0.35 &
      $\pm$0.06 &
      0.53 &
      $\pm$0.23 &
      0.33 &
      $\pm$0.00 &
      1.00 &
      $\pm$0.00 &
      1.00 &
      $\pm$0.00 \\ \midrule
    \multirow{4}{*}{\begin{tabular}[c]{@{}c@{}}Doors,\\ Keys \&\\ Gems\\ (3 Goals)\end{tabular}} &
      SIPS &
      0.35 &
      $\pm$0.07 &
      0.51 &
      $\pm$0.32 &
      0.54 &
      $\pm$0.37 &
      0.75 &
      $\pm$0.44 &
      0.75 &
      $\pm$0.44 &
      0.70 &
      $\pm$0.47 \\
     &
      BIRL-U &
      0.33 &
      $\pm$0.00 &
      0.33 &
      $\pm$0.00 &
      0.33 &
      $\pm$0.00 &
      0.33 &
      $\pm$0.00 &
      0.33 &
      $\pm$0.00 &
      0.33 &
      $\pm$0.00 \\
     &
      BIRL-O &
      0.34 &
      $\pm$0.02 &
      0.36 &
      $\pm$0.04 &
      0.43 &
      $\pm$0.07 &
      0.4 &
      $\pm$0.49 &
      0.55 &
      $\pm$0.50 &
      0.75 &
      $\pm$0.43 \\
     &
      PRP &
      0.35 &
      $\pm$0.17 &
      0.38 &
      $\pm$0.32 &
      0.64 &
      $\pm$0.40 &
      0.90 &
      $\pm$0.31 &
      0.70 &
      $\pm$0.47 &
      0.83 &
      $\pm$0.38 \\ \midrule
    \multirow{4}{*}{\begin{tabular}[c]{@{}c@{}}Block\\ Words\\ (5 Goals)\end{tabular}} &
      SIPS &
      0.52 &
      $\pm$0.33 &
      0.89 &
      $\pm$0.28 &
      0.96 &
      $\pm$0.18 &
      0.80 &
      $\pm$0.41 &
      0.90 &
      $\pm$0.31 &
      0.97 &
      $\pm$0.18 \\
     &
      BIRL-U &
      0.19 &
      $\pm$0.03 &
      0.19 &
      $\pm$0.03 &
      0.19 &
      $\pm$0.04 &
      0.37 &
      $\pm$0.48 &
      0.47 &
      $\pm$0.50 &
      0.53 &
      $\pm$0.50 \\
     &
      BIRL-O &
      0.19 &
      $\pm$0.03 &
      0.29 &
      $\pm$0.06 &
      0.45 &
      $\pm$0.09 &
      0.73 &
      $\pm$0.44 &
      0.77 &
      $\pm$0.42 &
      0.93 &
      $\pm$0.25 \\
     &
      PRP &
      0.36 &
      $\pm$0.18 &
      0.77 &
      $\pm$0.24 &
      0.91 &
      $\pm$0.17 &
      1.00 &
      $\pm$0.00 &
      1.00 &
      $\pm$0.00 &
      1.00 &
      $\pm$0.00 \\ \midrule
    \multirow{4}{*}{\begin{tabular}[c]{@{}c@{}}Intrusion\\ Detection\\ (20 Goals)\end{tabular}} &
      SIPS &
      0.52 &
      $\pm$0.43 &
      0.80 &
      $\pm$0.41 &
      0.80 &
      $\pm$0.41 &
      0.58 &
      $\pm$0.50 &
      0.80 &
      $\pm$0.41 &
      0.80 &
      $\pm$0.41 \\
     &
      BIRL-U &
      0.05 &
      $\pm$0.00 &
      0.05 &
      $\pm$0.00 &
      0.05 &
      $\pm$0.00 &
      0.05 &
      $\pm$0.00 &
      0.05 &
      $\pm$0.00 &
      0.05 &
      $\pm$0.00 \\
     &
      BIRL-O &
      0.09 &
      $\pm$0.01 &
      0.23 &
      $\pm$0.04 &
      0.52 &
      $\pm$0.07 &
      0.92 &
      $\pm$0.22 &
      1.00 &
      $\pm$0.00 &
      1.00 &
      $\pm$0.00 \\
     &
      PRP &
      0.42 &
      $\pm$0.01 &
      0.99 &
      $\pm$0.003 &
      1.00 &
      $\pm$0.00 &
      1.00 &
      $\pm$0.00 &
      1.00 &
      $\pm$0.00 &
      1.00 &
      $\pm$0.00 \\ \bottomrule
    \end{tabular}
    \vspace{6pt}
    \caption{Inference accuracy on the dataset of suboptimal trajectories.}
    \label{tab:accuracy-suboptimal-dataset}
\end{table}

\pagebreak

In terms of accuracy alone, it can be seen that the PRP baseline generally achieves the highest metrics, with SIPS and BIRL-O performing comparably, and with BIRL-U completely incapable of making accurate inferences except in the Taxi domain. As demonstrated by the qualitative comparisons however, these metrics alone maybe misleading, failing to show how inferences of each method really evolve over time. In particular, while the PRP baseline is routinely able to achieve the highest Top-1 accuracy, this may not correspond to a suitably calibrated posterior over goals, nor might it capture the sharp human-like changes over time that SIPS appears to display. It should also be noted that most of the domains considered do not allow for irreversible failures. As such, the distinctive capability of SIPS to infer goals despite failed plans is not captured by the results in Table \ref{tab:accuracy-suboptimal-dataset}.

\begin{table}[h]
\centering
\scriptsize
\begin{tabular}{@{}clrrrrrrrr@{}}
\toprule
\multicolumn{1}{l}{} &
   &
  \multicolumn{8}{c}{\textbf{Runtime}} \\ \midrule
\multicolumn{1}{c}{\textbf{Domain}} &
  \textbf{Method} &
  \multicolumn{2}{c}{C0 (s)} &
  \multicolumn{2}{c}{MC (s)} &
  \multicolumn{2}{c}{AC (s)} &
  \multicolumn{2}{c}{N} \\ \midrule
\multirow{4}{*}{\begin{tabular}[c]{@{}c@{}}Taxi\\ (3 Goals)\end{tabular}} &
  SIPS &
  14.7 &
  $\pm$6.73 &
  2.19 &
  $\pm$0.95 &
  3.08 &
  $\pm$1.24 &
  1220 &
  $\pm$405 \\
                     & BIRL-U & 2.22  & $\pm$0.06 & 0.002 & $\pm$0.0007 & 0.17 & $\pm$0.03 & 10000  & $\pm$0     \\
                     & BIRL-O & 0.56  & $\pm$0.02 & 0.002 & $\pm$0.0006 & 0.04 & $\pm$0.01 & 2500   & $\pm$0     \\
                     & PRP    & 13.2  & $\pm$2.19 & 6.21  & $\pm$1.52 & 6.73 & $\pm$1.50 & 6830   & $\pm$2090  \\ \midrule
\multirow{4}{*}{\begin{tabular}[c]{@{}c@{}}Doors,\\ Keys \& \\ Gems\\ (3 Goals)\end{tabular}} &
  SIPS &
  3.17 &
  $\pm$1.10 &
  0.72 &
  $\pm$0.21 &
  0.84 &
  $\pm$0.25 &
  2100 &
  $\pm$1140 \\
 & BIRL-U & 3280  & $\pm$173  & 0.13  & $\pm$0.14 & 181  & $\pm$184  & 250000 & $\pm$0     \\
 & BIRL-O & 142   & $\pm$13.0 & 0.13  & $\pm$0.14 & 8.00 & $\pm$8.24 & 10000  & $\pm$0     \\
 & PRP    & 5.32  & $\pm$2.21 & 3.12  & $\pm$1.58 & 3.24 & $\pm$1.67 & 5970   & $\pm$3350  \\ \midrule
\multirow{4}{*}{\begin{tabular}[c]{@{}c@{}}Block\\ Words\\ (5 Goals)\end{tabular}} &
  SIPS &
  21.1 &
  $\pm$4.84 &
  1.67 &
  $\pm$0.61 &
  3.62 &
  $\pm$0.85 &
  2380 &
  $\pm$1110 \\
                     & BIRL-U & 687   & $\pm$273  & 0.15  & $\pm$0.05 & 69.5 & $\pm$31.2 & 250000 & $\pm$0     \\
                     & BIRL-O & 19.5  & $\pm$0.59 & 0.12  & $\pm$0.03 & 2.11 & $\pm$0.51 & 10000  & $\pm$0     \\
                     & PRP    & 25.6  & $\pm$11.3 & 26.5  & $\pm$7.90 & 26.3 & $\pm$7.50 & 3980   & $\pm$1410  \\ \midrule
\multirow{4}{*}{\begin{tabular}[c]{@{}c@{}}Intrusion\\ Detection\\ (20 Goals)\end{tabular}} &
  SIPS &
  325 &
  $\pm$24.9 &
  12.0 &
  $\pm$1.40 &
  30.0 &
  $\pm$3.00 &
  14100 &
  $\pm$343 \\
                     & BIRL-U & 18000 & $\pm$2050 & 0.01  & $\pm$0.07 & 1130 & $\pm$230  & 250000 & $\pm$0     \\
                     & BIRL-O & 100   & $\pm$11.7 & 0.02  & $\pm$0.00 & 5.80 & $\pm$0.86 & 10000  & $\pm$0     \\
                     & PRP    & 246   & $\pm$5.12 & 381   & $\pm$108  & 374  & $\pm$102  & 75700  & $\pm$20800 \\ \bottomrule
\end{tabular}
\vspace{6pt}
\caption{Inference runtime on the dataset of optimal trajectories.}
\label{tab:runtime-optimal-dataset}
\end{table}

\begin{table}[h]
    \centering
    \scriptsize
    \begin{tabular}{clllllllll}
    \toprule
    \multicolumn{1}{l}{} &
       &
      \multicolumn{8}{c}{\textbf{Runtime}} \\ \midrule
    \textbf{Domain} &
      \multicolumn{1}{c}{\textbf{Method}} &
      \multicolumn{2}{c}{C0 (s)} &
      \multicolumn{2}{c}{MC (s)} &
      \multicolumn{2}{c}{AC (s)} &
      \multicolumn{2}{c}{N} \\ \midrule
    \multirow{4}{*}{\begin{tabular}[c]{@{}c@{}}Taxi\\ (3 Goals)\end{tabular}} &
      SIPS &
      12.2 &
      $\pm$7.75 &
      1.61 &
      $\pm$0.74 &
      2.29 &
      $\pm$1.05 &
      1530 &
      $\pm$1110 \\
     &
      BIRL-U &
      2.22 &
      $\pm$0.06 &
      0.003 &
      $\pm$0.0004 &
      0.16 &
      $\pm$0.04 &
      10000 &
      $\pm$0.00 \\
     &
      BIRL-O &
      2.17 &
      $\pm$0.05 &
      0.002 &
      $\pm$0.0003 &
      0.15 &
      $\pm$0.04 &
      2500 &
      $\pm$0.00 \\
     &
      PRP &
      13.3 &
      $\pm$3.26 &
      7.33 &
      $\pm$2.61 &
      7.74 &
      $\pm$2.56 &
      8840 &
      $\pm$5800 \\ \midrule
    \multirow{4}{*}{\begin{tabular}[c]{@{}c@{}}Doors,\\ Keys \&\\ Gems\\ (3 Goals)\end{tabular}} &
      SIPS &
      3.40 &
      $\pm$1.18 &
      0.69 &
      $\pm$0.24 &
      0.87 &
      $\pm$0.31 &
      2100 &
      $\pm$1140 \\
     &
      BIRL-U &
      3360 &
      $\pm$66.0 &
      0.11 &
      $\pm$0.06 &
      133 &
      $\pm$68.7 &
      250000 &
      $\pm$0.00 \\
     &
      BIRL-O &
      155 &
      $\pm$3.31 &
      0.11 &
      $\pm$0.06 &
      6.27 &
      $\pm$3.31 &
      10000 &
      $\pm$0.00 \\
     &
      PRP &
      4.65 &
      $\pm$1.58 &
      3.04 &
      $\pm$1.56 &
      3.11 &
      $\pm$1.56 &
      6150 &
      $\pm$3680 \\ \midrule
    \multirow{4}{*}{\begin{tabular}[c]{@{}c@{}}Block\\ Words\\ (5 Goals)\end{tabular}} &
      SIPS &
      20.6 &
      $\pm$5.79 &
      2.86 &
      $\pm$1.12 &
      4.41 &
      $\pm$1.77 &
      2570 &
      $\pm$810 \\
     &
      BIRL-U &
      687 &
      $\pm$273 &
      0.33 &
      $\pm$0.13 &
      60.6 &
      $\pm$34.0 &
      250000 &
      $\pm$0.00 \\
     &
      BIRL-O &
      23.5 &
      $\pm$1.76 &
      0.01 &
      $\pm$0.001 &
      2.12 &
      $\pm$0.86 &
      10000 &
      $\pm$0.00 \\
     &
      PRP &
      40.5 &
      $\pm$22.7 &
      38.9 &
      $\pm$16.1 &
      38.9 &
      $\pm$15.7 &
      5660 &
      $\pm$4860 \\ \midrule
    \multirow{4}{*}{\begin{tabular}[c]{@{}c@{}}Intrusion\\ Detection\\ (20 Goals)\end{tabular}} &
      SIPS &
      400 &
      $\pm$29.7 &
      3.90 &
      $\pm$1.04 &
      26.6 &
      $\pm$2.06 &
      12900 &
      $\pm$3020 \\
     &
      BIRL-U &
      18000 &
      $\pm$2050 &
      1.12 &
      $\pm$3.83 &
      1040 &
      $\pm$163 &
      250000 &
      $\pm$0.00 \\
     &
      BIRL-O &
      96.9 &
      $\pm$10.4 &
      0.02 &
      $\pm$0.002 &
      5.60 &
      $\pm$0.77 &
      10000 &
      $\pm$0.00 \\
     &
      PRP &
      281 &
      $\pm$2.48 &
      332 &
      $\pm$25.8 &
      330 &
      $\pm$24.7 &
      51900 &
      $\pm$960 \\ \midrule
    \end{tabular}
    \vspace{6pt}
    \caption{Inference runtime on the dataset of suboptimal trajectories.}
    \label{tab:runtime-suboptimal-dataset}
\end{table}

Once runtime is taken into account, it becomes clear that SIPS achieves the best balance between speed and accuracy due to its use of incremental computation. In contrast, BIRL-U requires orders of magnitude more initial computation while still failing to produce meaningful inferences, while PRP requires up to 30 times more computation per timestep. This is especially apparent on the Intrusion Detection domain, which has a large number of goals, requiring PRP to compute a large number of optimal plans at each timestep. Even the BIRL-O baseline, which assumes oracular access to the dataset of observed trajectories during value iteration, is slower than SIPS on the Doors, Keys \& Gems domain in terms of average runtime. Overall, these results imply that SIPS is the only method suitable for online usage on the full range of domains we consider.

\subsection{Robustness to Parameter Mismatch}

Tables \ref{tab:robustness-doorskeysgems} and \ref{tab:robustness-blockwords} present additional results for the robustness experiments described in the main text, showing how different settings of model parameters fare against each other. Each column corresponds to a parameter value assumed by SIPS, and each row corresponds to the true parameter for the boundedly rational agent model used to generate the data. Within each sub-table, unspecified parameters default to $\gamma=0.1$, $r=2$, $q=0.95$, $h=$Manhattan (for Doors, Keys, Gems) and $h=h_\text{add}$ (for Block Words).

It can be seen that SIPS fares reasonably well against mismatched parameters, with degradation partly driven by mismatch itself, but also partly by increased randomness when the data-generating parameters lead to less optimal agent behavior. The effect of noisy behavior is especially apparent in Table \ref{tab:robustness-blockwords}(d): data generated by agents using the highly uninformative goal count heuristic (which simply the counts the number of goal predicates yet to be satisfied as a distance metric) is highly random. This results in very poor inferences (Top-1 at Q3 = 0.37), even when SIPS correctly assumes the same heuristic. Nonetheless, mismatched heuristics do lead to poorer performance, raising the open question of whether observers need good models of others' planning heuristics in order to accurately infer their goals.

\begin{table}[h]
\centering
\tiny
\begin{subtable}[b]{0.225\textwidth}
\begin{tabular}{@{}lllll@{}}
\toprule
 &  & \multicolumn{3}{c}{\textbf{Assumed}} \\
 & $r$ & 1 & 2 & 4 \\ \midrule
\multirow{3}{*}{\rotatebox[origin=c]{90}{\textbf{True}}} & 1 & 0.80 & 0.60 & 0.70 \\
 & 2 & 0.73 & 0.73 & 0.77 \\
 & 4 & 0.77 & 0.73 & 0.87 \\ \bottomrule
\end{tabular}
\caption{Persistence ($r$)}
\end{subtable}
\begin{subtable}[b]{0.24\textwidth}
\begin{tabular}{@{}lllll@{}}
\toprule
 &  & \multicolumn{3}{c}{\textbf{Assumed}} \\
 & $q$ & 0.80 & 0.90 & 0.95 \\ \midrule
\multirow{3}{*}{\rotatebox[origin=c]{90}{\textbf{True}}} & 0.80 & 0.73 & 0.67 & 0.53 \\
 & 0.90 & 0.77 & 0.63 & 0.60 \\
 & 0.95 & 0.77 & 0.60 & 0.73 \\ \bottomrule
\end{tabular}
\caption{Persistence ($q$)}
\end{subtable}
\begin{subtable}[b]{0.24\textwidth}
\begin{tabular}{@{}lllll@{}}
\toprule
 &  & \multicolumn{3}{c}{\textbf{Assumed}} \\
 & $\gamma$ & 0.02 & 0.10 & 0.50 \\ \midrule
\multirow{3}{*}{\rotatebox[origin=c]{90}{\textbf{True}}} & 0.02 & 0.90 & 0.77 & 0.83 \\
 & 0.10 & 0.77 & 0.73 & 0.73 \\
 & 0.50 & 0.73 & 0.67 & 0.77 \\ \bottomrule
\end{tabular}
\caption{Search noise ($\gamma$)}
\end{subtable}
\begin{subtable}[b]{0.24\textwidth}
\begin{tabular}{@{}lllll@{}}
\toprule
 &  & \multicolumn{3}{c}{\textbf{Assumed}} \\
 & $h$ & Mh. &  & Mz. \\ \midrule
\multirow{3}{*}{\rotatebox[origin=c]{90}{\textbf{True}}} & Mh. & 0.83 &  & 0.77 \\
 &  &  &  &  \\
 & Mz. & 0.80 &  & 0.90 \\ \bottomrule
\end{tabular}
\caption{Heuristic ($h$)}
\end{subtable}
\caption{Robustness to parameter mismatch for the Doors, Keys, Gems domain. The metric shown is the top-1 accuracy of SIPS at the third time quartile (Q3). $h$=Mh. refers to Manhattan distance, while $h$=Mz. refers to maze distance.}
\label{tab:robustness-doorskeysgems}
\end{table}

\begin{table}[h]
\centering
\tiny
\begin{subtable}[b]{0.225\textwidth}
\begin{tabular}{@{}lllll@{}}
\toprule
 &  & \multicolumn{3}{c}{\textbf{Assumed}} \\
 & $r$ & 1 & 2 & 4 \\ \midrule
\multirow{3}{*}{\rotatebox[origin=c]{90}{\textbf{True}}} & 1 & 0.80 & 0.90 & 0.77 \\
 & 2 & 0.83 & 0.80 & 0.90 \\
 & 4 & 0.87 & 0.90 & 0.93 \\ \bottomrule
\end{tabular}
\caption{Persistence ($r$)}
\end{subtable}
\begin{subtable}[b]{0.24\textwidth}
\begin{tabular}{@{}lllll@{}}
\toprule
 &  & \multicolumn{3}{c}{\textbf{Assumed}} \\
 & $q$ & 0.80 & 0.90 & 0.95 \\ \midrule
\multirow{3}{*}{\rotatebox[origin=c]{90}{\textbf{True}}} & 0.80 & 0.80 & 0.83 & 0.70 \\
 & 0.90 & 0.70 & 0.80 & 0.83 \\
 & 0.95 & 0.83 & 0.80 & 0.87 \\ \bottomrule
\end{tabular}
\caption{Persistence ($q$)}
\end{subtable}
\begin{subtable}[b]{0.24\textwidth}
\begin{tabular}{@{}lllll@{}}
\toprule
 &  & \multicolumn{3}{c}{\textbf{Assumed}} \\
 & $\gamma$ & 0.02 & 0.10 & 0.50 \\ \midrule
\multirow{3}{*}{\rotatebox[origin=c]{90}{\textbf{True}}} & 0.02 & 0.80 & 0.87 & 0.87 \\
 & 0.10 & 0.83 & 0.87 & 0.83 \\
 & 0.50 & 0.90 & 0.83 & 0.87 \\ \bottomrule
\end{tabular}
\caption{Search noise ($\gamma$)}
\end{subtable}
\begin{subtable}[b]{0.24\textwidth}
\begin{tabular}{@{}lllll@{}}
\toprule
 &  & \multicolumn{3}{c}{\textbf{Assumed}} \\
 & $h$ & GC &  & $h_{\text{add}}$ \\ \midrule
\multirow{3}{*}{\rotatebox[origin=c]{90}{\textbf{True}}} & GC. & 0.37 &  & 0.43 \\
 &  &  &  &  \\
 & $h_{\text{add}}$ & 0.33 &  & 0.77 \\ \bottomrule
\end{tabular}
\caption{Heuristic ($h$)}
\end{subtable}
\caption{Robustness to parameter mismatch for the Blocks World domain. The metric shown is the top-1 accuracy of SIPS at the third time quartile (Q3). $h$=GC. refers to the goal count heuristic, while $h=h_\text{add}$ refers to the additive delete-relaxation heuristic.}
\label{tab:robustness-blockwords}
\end{table}

\section{Human Studies}

As described in the main text, we conducted two sets of pilot studies with human subjects, the first to measure human goal inferences for comparison, and the second to collect human-generated plans for robustness experiments. These studies were approved under MIT's IRB (COUHES no.: 0812003014).

\subsection{Human Inferences}

\begin{figure}[h]
    \centering
    \includegraphics[width=\textwidth]{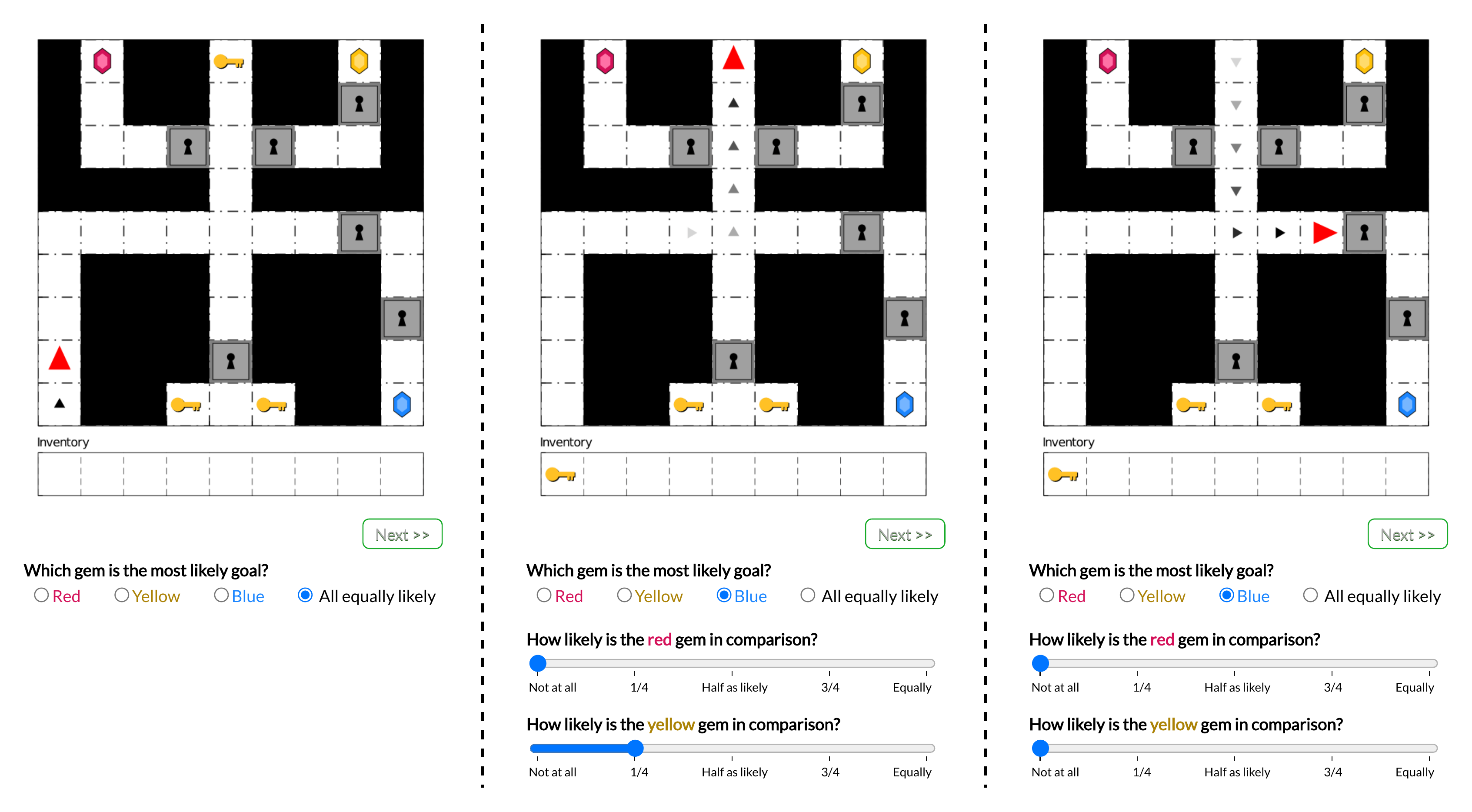}
    \caption{Web interface for collecting human goal inferences. Each panel shows one step in a sequence of judgement points presented to a participant.}
    \label{fig:experiment-interface}
\end{figure}

Data was collected from $N$=5 pilot subjects in the MIT population. Each subject was given access to a web interface that would present trajectories of an agent in the Doors, Keys \& Gems domain, and that would ask for goal inference judgements at every 6th timestep, as well as the first and last timestep. Subjects could select which gem they believed to be the most likely goal of the agent, and then were allowed to adjust sliders indicating how likely the other goals were in comparison. These relative probability ratings were normalized, and recorded. Excerpts from this interface are shown in Figure \ref{fig:experiment-interface}. Subjects were shown a series of 10 trajectories, out of which 4 were optimal trajectories, and 6 exhibited notable suboptimality or failure.

\pagebreak

\subsection{Human Plans}

Data was collected from $N$=5 pilot subjects in the MIT population. Each subject received a experimental script to run, which collected data for both the Doors-Keys-Gems domain and the Blocks World domain. For the Blocks World domain, data was collected for all combinations of 3 problems with 5 possible goals, and for the Doors-Keys-Gems domain, data was collected for all combinations of 5 problems with 3 possible goals.

For each pair consisting of a problem and a goal, the subjects were presented with a visualization of the initial state and a textual description of their goal. The subjects were then presented with a list of keys corresponding to the actions available from the current state, and prompted to press the key corresponding to their selected action. Once the subjects entered their action of choice, the visualization would update to show the state after the action had occurred. The subjects would then be prompted again for an action. This process repeated until the given goal was achieved, or the subject terminated that task (e.g. if the goal was no longer achievable). Once the goal was achieved for a given problem and goal pair, the sequence of actions was recorded.